\documentclass{IEEEoj-data}
\usepackage{cite}
\usepackage{amsmath,amssymb,amsfonts}
\usepackage{algorithmic}
\usepackage{graphicx,color}
\usepackage{textcomp}
\usepackage{hyperref}
\usepackage{balance}
\usepackage{soul}
\def\BibTeX{{\rm B\kern-.05em{\sc i\kern-.025em b}\kern-.08em
    T\kern-.1667em\lower.7ex\hbox{E}\kern-.125emX}}
\AtBeginDocument{\definecolor{ojcolor}{cmyk}{0.93,0.59,0.15,0.02}}

\begin{document}
\receiveddate{20 February 2026}
\reviseddate{31 March 2026 and 09 April 2026}
\accepteddate{09 April 2026}
\doiinfo{10.1109/IEEEDATA.2026.3683381}

\title{\textcolor{black}{Descriptor:} \textcolor{ieeedata}{\textit{Parasitoid Wasps and Associated Hymenoptera Dataset}} (DAPWH)}

\author{
    JOÃO MANOEL HERRERA PINHEIRO\authorrefmark{1} (STUDENT MEMBER, IEEE), 
    GABRIELA DO NASCIMENTO HERRERA\authorrefmark{2}, 
    LUCIANA BUENO DOS REIS FERNANDES\authorrefmark{2},
    ALVARO DORIA DOS SANTOS\authorrefmark{3},
    RICARDO V. GODOY\authorrefmark{1},
    EDUARDO A. B. ALMEIDA\authorrefmark{4}, 
    HELENA CAROLINA ONODY\authorrefmark{5},
    MARCELO ANDRADE DA COSTA VIEIRA\authorrefmark{1},
    ANGÉLICA MARIA PENTEADO-DIAS\authorrefmark{2} AND
    MARCELO BECKER\authorrefmark{1} (MEMBER, IEEE)
}

\affil{São Carlos School of Engineering, University of São Paulo, São Carlos 13566590, SP, Brazil}
\affil{Department of Ecology and Evolutionary Biology, Federal University of São Carlos, São Carlos 13565905, Brazil}
\affil{Federal University of Tocantins, Porto Nacional, 77500000, Brazil}
\affil{Department of Biology (FFCLRP), University of São Paulo, Ribeirão Preto 14040901, Brazil}
\affil{State University of Piauí, Deputado Jesualdo Cavalcanti Campus, Corrente, 49800000, Brazil}

\corresp{CORRESPONDING AUTHOR: João Manoel Herrera Pinheiro (e-mail: joao.manoel.pinheiro@usp.br), Gabriela do Nascimento Herrera (e-mail: gabriela.herrera@estudante.ufscar.br
), Angélica Maria Pentado-Dias (e-mail: angelica@ufscar.br) and Marcelo Becker (e-mail: becker@sc.usp.br).}
\markboth{IEEE-DATA  PARASITOID WASPS AND ASSOCIATED HYMENOPTERA DATASET}{PINHEIRO \textit{et al.}}
\authornote{This article has been accepted for publication in IEEE Data Descriptions. This is the author’s version which has not been fully edited and content may change prior to final publication. Citation information: DOI 10.1109/IEEEDATA.2026.3683381}
\begin{abstract}
Accurate taxonomic identification is the cornerstone of biodiversity monitoring and agricultural management, particularly for the hyper-diverse superfamily Ichneumonoidea. Comprising the families Ichneumonidae and Braconidae, these parasitoid wasps are ecologically critical for regulating insect populations, yet they remain one of the most taxonomically challenging groups due to their cryptic morphology and vast number of undescribed species. To address the scarcity of robust digital resources for these key groups, we present a curated image dataset designed to advance automated identification systems. The dataset contains 3,556 high-resolution images, primarily focused on Neotropical Ichneumonidae and Braconidae, while also including supplementary families such as Andrenidae, Apidae, Bethylidae, Chrysididae, Colletidae, Halictidae, Megachilidae, Pompilidae, and Vespidae to improve model robustness. Crucially, a subset of 1,739 images is annotated in COCO format, featuring multi-class bounding boxes for the full insect body, wing venation, and scale bars. This resource provides a foundation for developing computer vision models capable of identifying these families.
 \\ 
 {\textcolor{ieeedata}{\abstractheadfont\bfseries{IEEE SOCIETY/COUNCIL}}}     IEEE Computational Intelligence Society (CIS)\\  
 \\
 {\textcolor{ieeedata}{\abstractheadfont\bfseries{DATA DOI/PID}}}     \hyperlink{https://www.doi.org/10.5281/zenodo.18501018}{10.5281/zenodo.18501018}\\ 
  
 {\textcolor{ieeedata}{\abstractheadfont\bfseries{DATA TYPE/LOCATION}}}  Images, Brazil

\end{abstract}


\maketitle

\section*{BACKGROUND} 

The order Hymenoptera, which includes bees, wasps, and ants \cite{HuberHym2017}, is among the most speciose groups within Hexapoda \cite{2015Nigel,2018Nigel}. Although Coleoptera and Lepidoptera presently account for the greatest number of described species, these four orders are widely recognized as the most species-rich overall, with accumulating evidence suggesting that Hymenoptera may in fact represent the most speciose order \cite{Forbes2018,2018Nigel,Eggleton2020}. The bees are the most famous insect family in the Hymenoptera order, they are recognized for their essential pollination services \cite{Ollerton2011}, while the population status of many other Hymenopterans, such as ants, wasps, and parasitoids, which also provide important ecosystem services, largely remains unknown \cite{SANCHEZBAYO2019}. Many species within this order serve as pollen vectors, contributing to the pollination of various plants and acting as effective pollinators \cite{Barbizan2009,Beggs2011}.

Parasitoid wasps are likely the most diverse group within the order Hymenoptera, potentially representing the largest group in terms of species numbers \cite{gauldHymenoptera1988,Quicke2015,Shaw2018,QuickeAsia2023,POLASZEK2023}. The Ichneumonoidea superfamily, represents a vast group of parasitoid wasps within the order Hymenoptera, these insects play a critical role in regulating the populations of other insect species~\cite{Scatolini2003,Schoeninger2019,Antunes2020}. As parasitoid Hymenoptera, they primarily attack the larvae and pupae of Lepidoptera and Coleoptera, many of which are important agricultural pests\cite{Antunes2020}. Beyond pest regulation, they also act as pollinators and hold significant importance in maintaining ecological diversity \cite{Sharkey1993,Sharkey2006}. With over 45,000 validly described species, a significant number of species within these groups remain undescribed or are challenging to identify~\cite{Quicke2015,HuberHym2017,Chen2019,insects13020170,POLASZEK2023}, highlighting a critical need for robust resources to aid their taxonomic.

\begin{figure}[htp]
    \centering
    \includegraphics[width=0.45\textwidth]{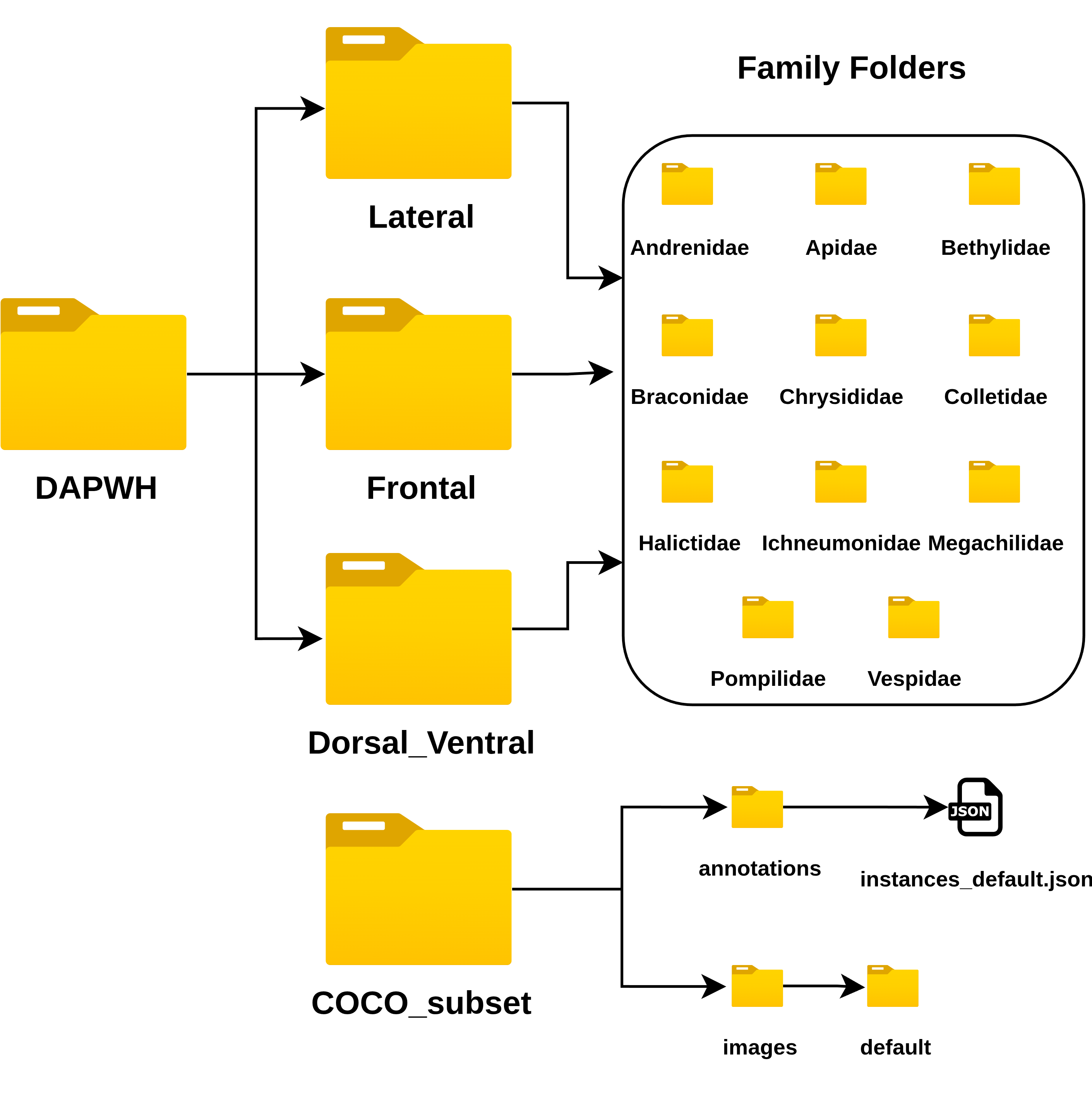} 
        \caption{Schematic representation of the dataset file organization. The primary repository (DAPWH) \cite{herrera_pinheiro_2026_18501018} is structured hierarchically by anatomical view (\textit{Lateral}, \textit{Frontal}, \textit{Dorsal\_Ventral}), with each view containing specific families' subdirectories. Complementing this structure, the \textbf{COCO\_subset} provides a subset version of the dataset, including the \texttt{instances\_default} json file and corresponding images.}
    \label{fig:flow_final_root}
\end{figure}

Traditional insect identification relies on expert taxonomists using microscopes and morphological keys, a slow and highly specialized process in a time of global taxonomist shortage \cite{Luke2023}. In response to the critical need for robust identification resources, we introduce a comprehensive dataset of 3,556 high-quality images centered on the parasitoid families Ichneumonidae and Braconidae. To ensure taxonomic robustness, the dataset includes supplementary images from diverse Hymenoptera groups, including Apidae, Vespidae, and Pompilidae, among others. A key contribution of this work is the inclusion of a fully annotated subset in COCO format; these annotations delineate the full body, wing structures, and scale bars, explicitly designed to train object detection architectures. Figure \ref{fig:flow_final_root} shows the schematic representation of the dataset organization.

It offers a valuable resource for researchers across disciplines, including entomologists, ecologists, and computer vision scientists, enabling the training and validation of machine learning and deep learning models for automated insect identification and fine-grained classification, particularly within taxonomically complex groups like Ichneumonoidea.

\section*{COLLECTION METHODS AND DESIGN} 

The dataset comprises high-resolution images of specimens from the superfamily Ichneumonoidea as well as nine additional hymenopteran families: Andrenidae, Apidae, Bethylidae, Chrysididae, Colletidae, Halictidae, Megachilidae, Pompilidae, and Vespidae. Specimens were obtained from four primary university affiliated repositories, the specimens curated in these repositories represent the culmination of years of research projects.

Originally sampled using Malaise traps during extensive field campaigns, the material was subsequently transferred to university laboratories. There, expert taxonomists identified the specimens using morphological dichotomous keys before permanently depositing them in the collections. All images are licensed under CC BY 4.0 or CC BY-NC 4.0.

\begin{figure}[htp]
    \centering
    \includegraphics[width=0.45\textwidth]{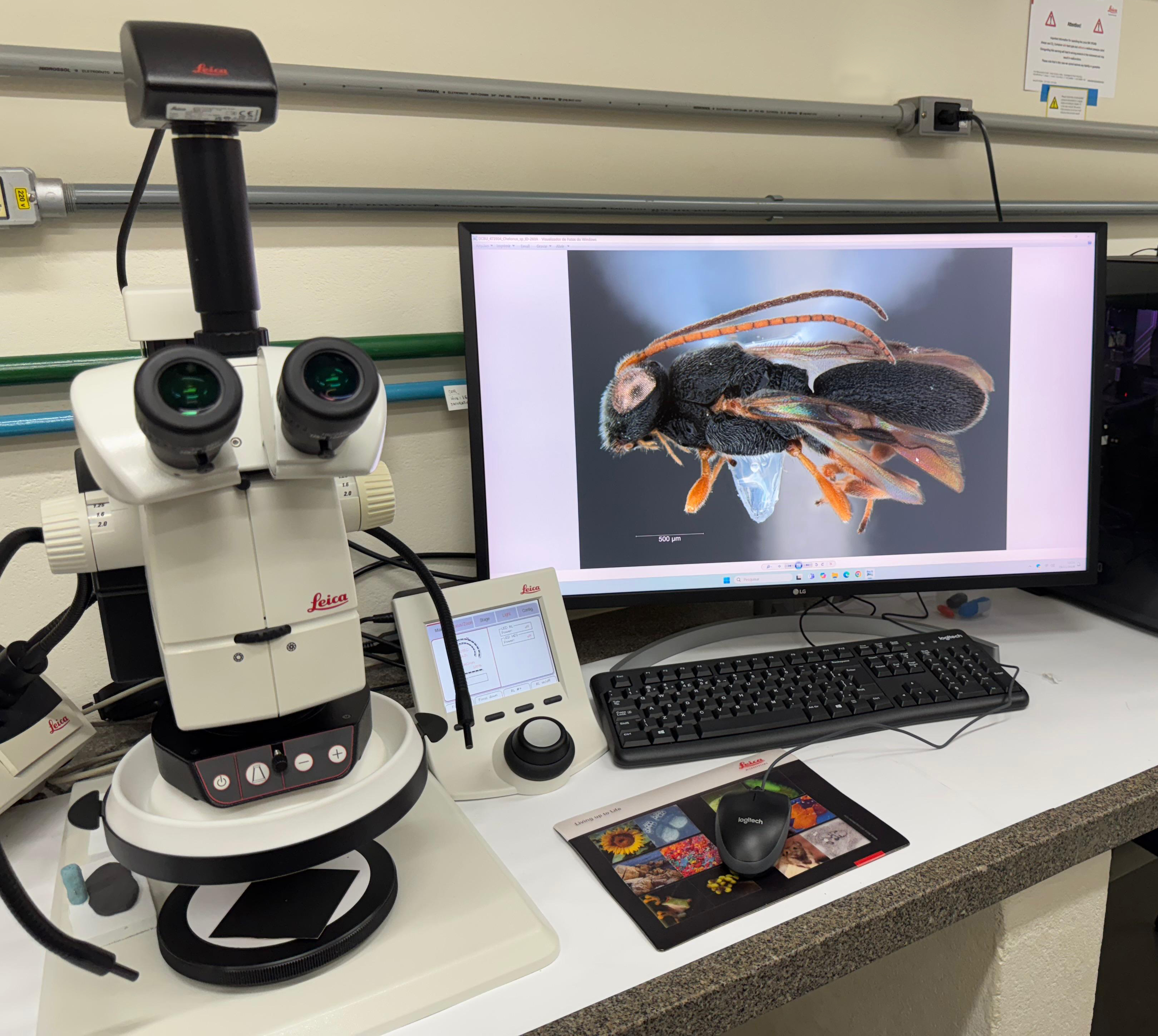} 
        \caption{Specimens were retrieved from DCBU collection, mounted under a Leica M205C stereomicroscope equipped with a Leica K5C camera.}
    \label{fig:photos}
\end{figure}

Specimens representing the families Braconidae and Ichneumonidae were primarily acquired from the Coleção Taxonômica do Departamento de Ecologia e Biologia Evolutiva da UFSCar (DCBU) \cite{dias2025dcbu}. The process consisted of retrieving specimens from the collection, mounting them under a Leica M205C stereomicroscope equipped with a Leica K5C camera was used for image acquisition, with LAS X software controlling capture. Image stacking was performed using Helicon Focus software. Figure \ref{fig:photos} illustrates the procedure adopted for specimen imaging.

\begin{figure}[htp]
    \centering
    \includegraphics[width=0.46\textwidth]{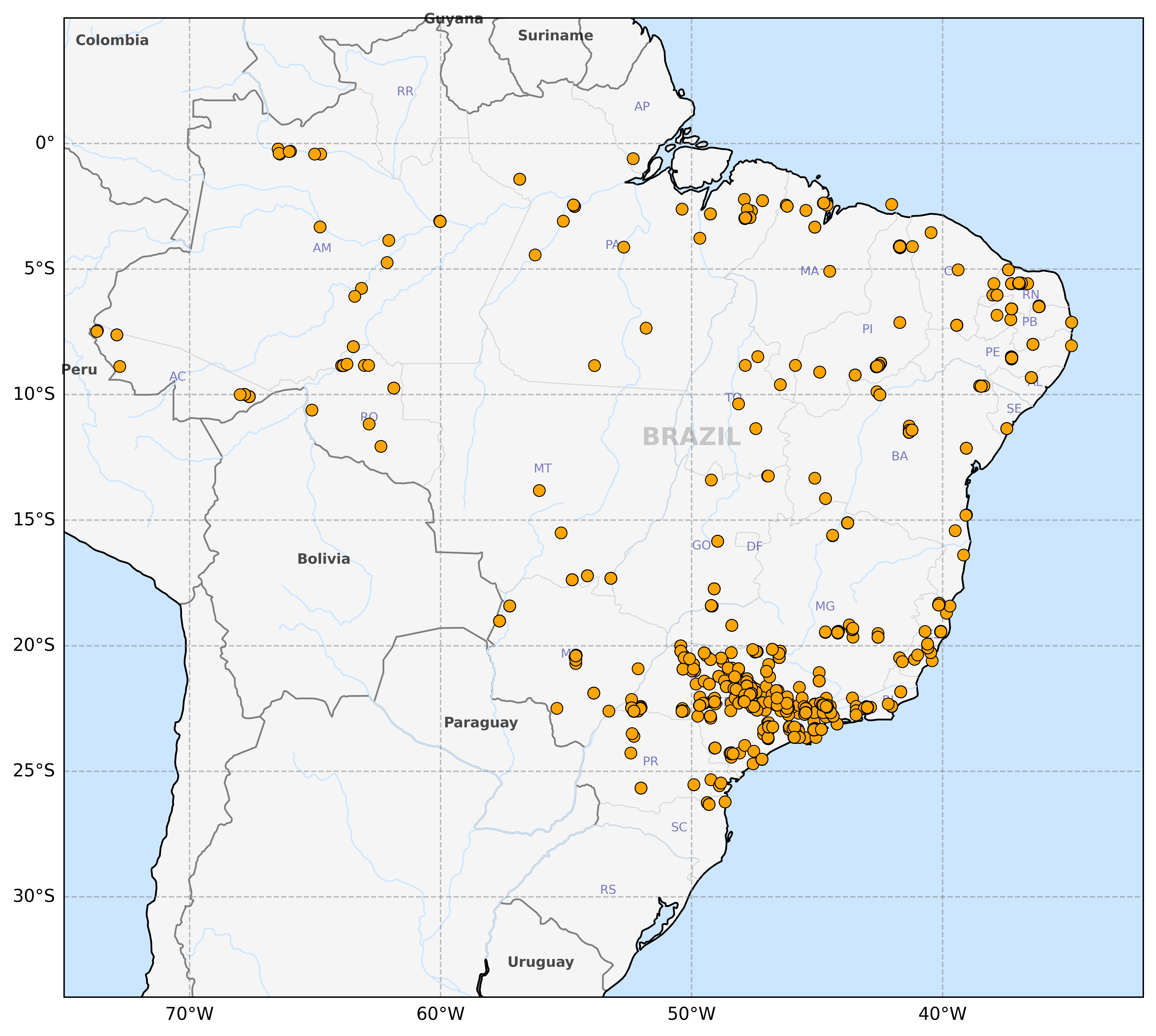} 
       \caption{Illustrative geographic distribution of specimen sources from the DCBU, MZUSP, and RPSP collections, generated using Python, Pandas, and Cartopy.}
    \label{fig:cap5_map_distribution}
\end{figure}

Figure \ref{fig:cap5_map_distribution} maps the collection localities of samples. The spatial distribution highlights that the vast majority of specimens originate from the Neotropical region, specifically Brazil.

\begin{figure*}[htp]
    \centering
    \begin{minipage}[t]{0.3\textwidth}
        \centering
        \includegraphics[width=\linewidth]{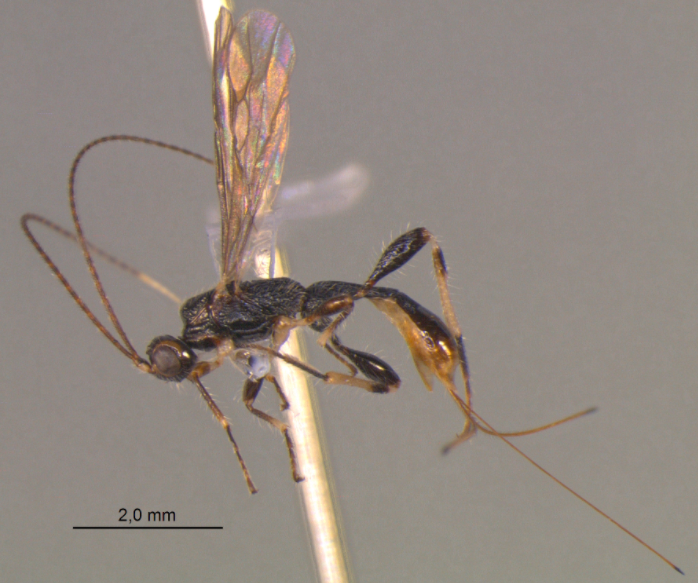}
        \par (a)
    \end{minipage}
    \hfill
    \begin{minipage}[t]{0.2\textwidth}
        \centering
        \includegraphics[width=\linewidth]{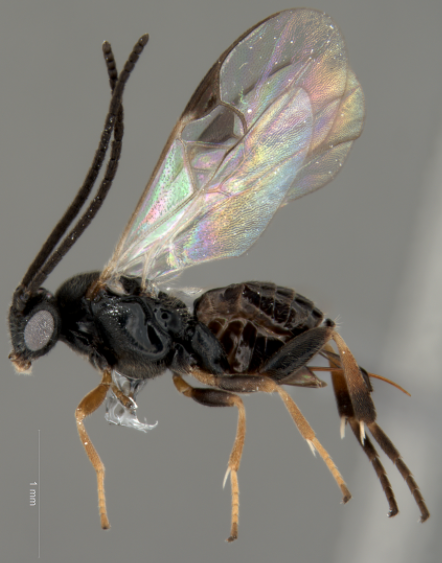}
        \par (b)
    \end{minipage}
    \hfill
        \begin{minipage}[t]{0.35\textwidth}
        \centering
        \includegraphics[width=\linewidth]{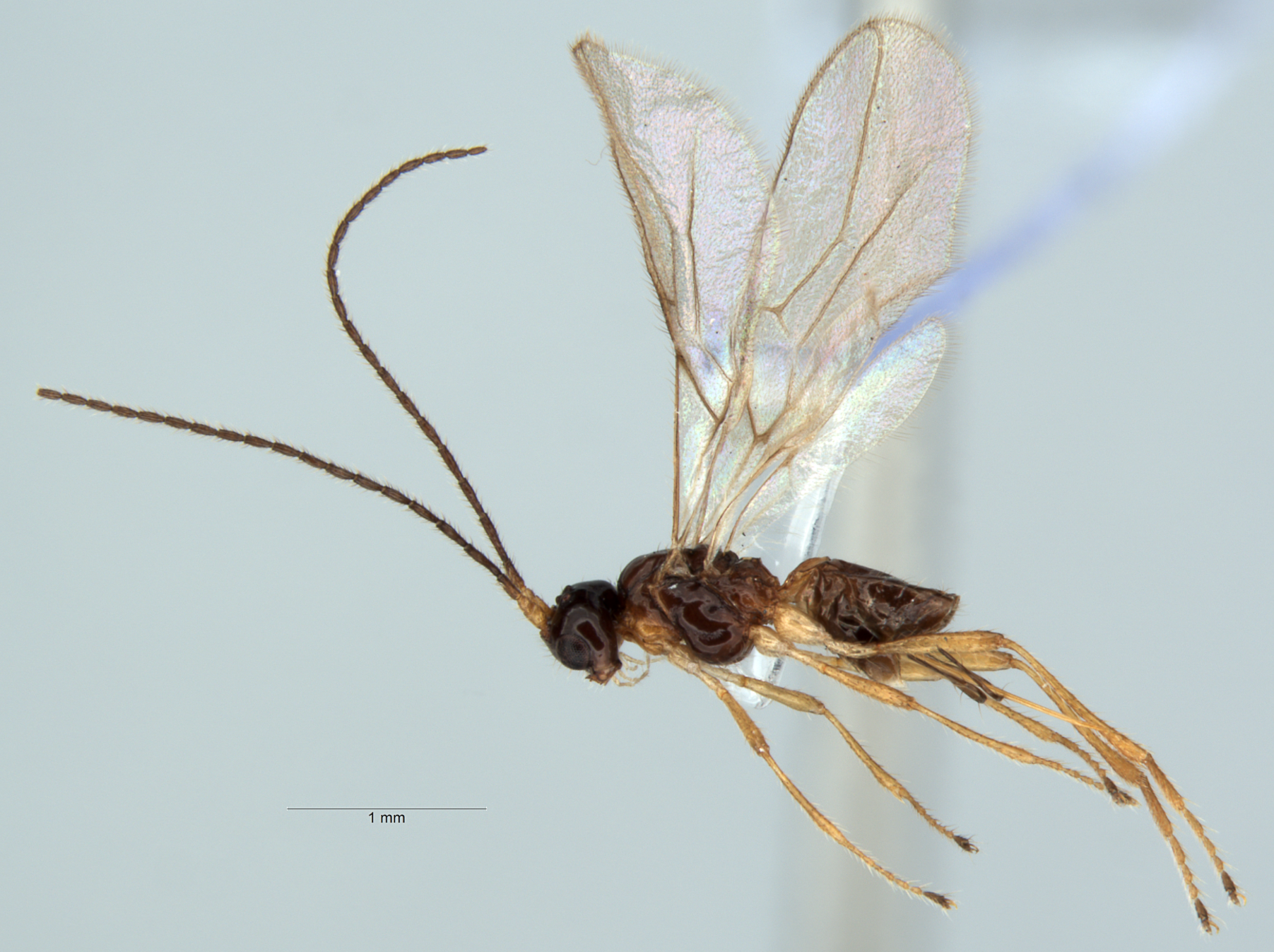}
        \par (c)
    \end{minipage}
    \vspace{0.5cm} 

    \begin{minipage}[t]{0.3\textwidth}
        \centering
        \includegraphics[width=\linewidth]{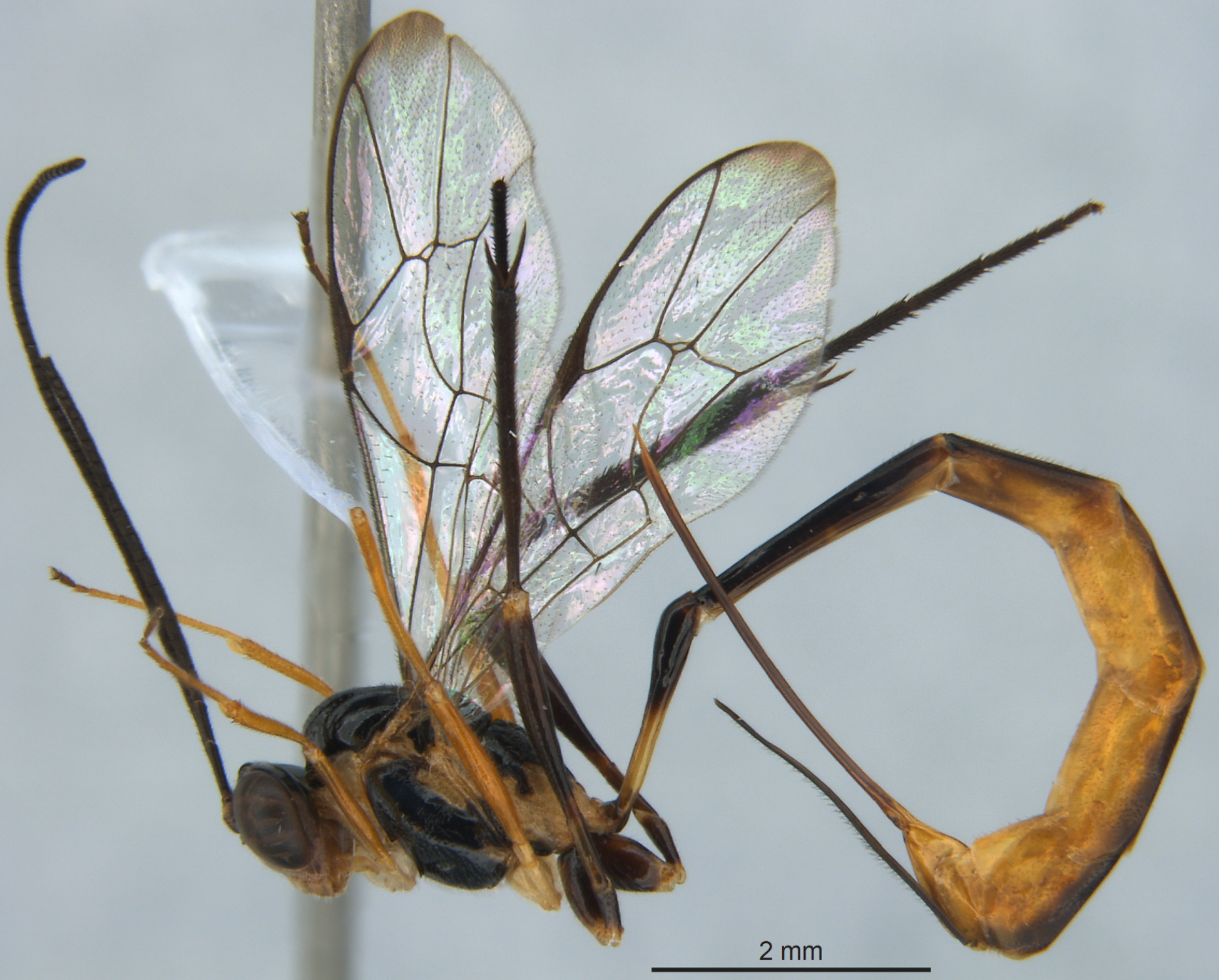}
        \par (d)
    \end{minipage}
    \hfill
        \begin{minipage}[t]{0.25\textwidth}
        \centering
        \includegraphics[width=\linewidth]{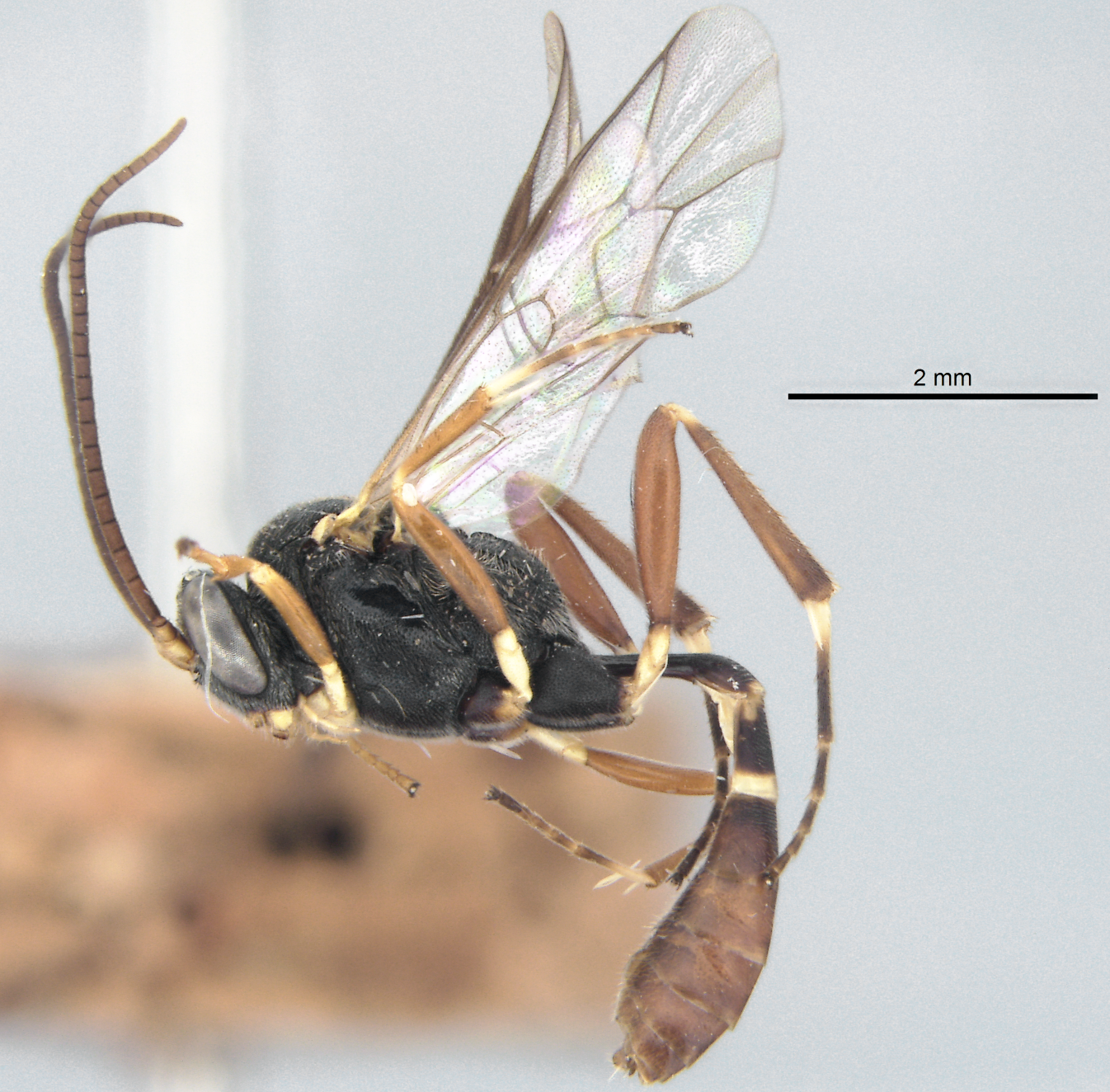}
        \par (e)
    \end{minipage}
    \hfill
        \hfill
        \begin{minipage}[t]{0.35\textwidth}
        \centering
        \includegraphics[width=\linewidth]{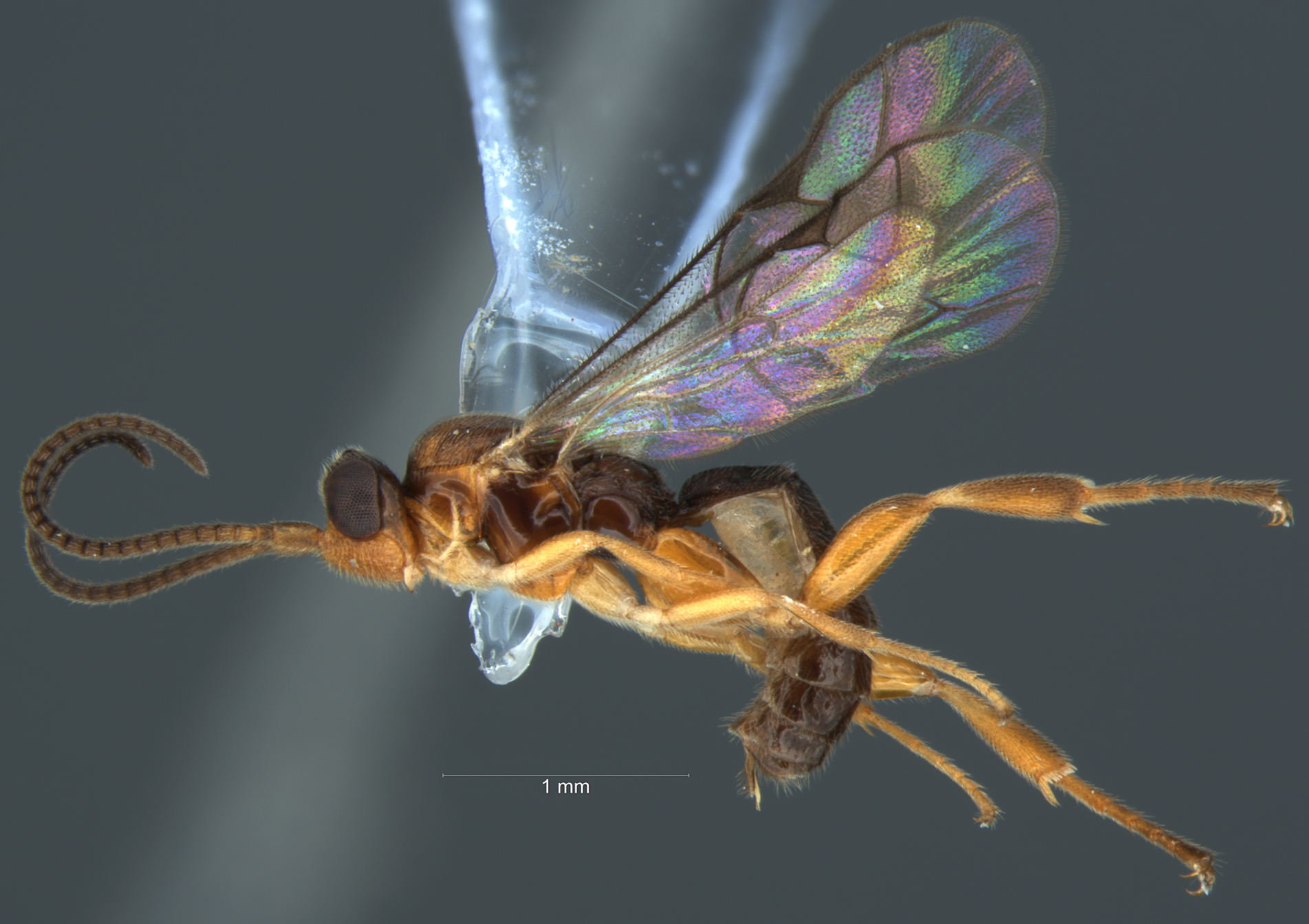}
        \par (f)
    \end{minipage}

    \vspace{0.5cm} 

    \begin{minipage}[t]{0.25\textwidth}
        \centering
        \includegraphics[width=\linewidth]{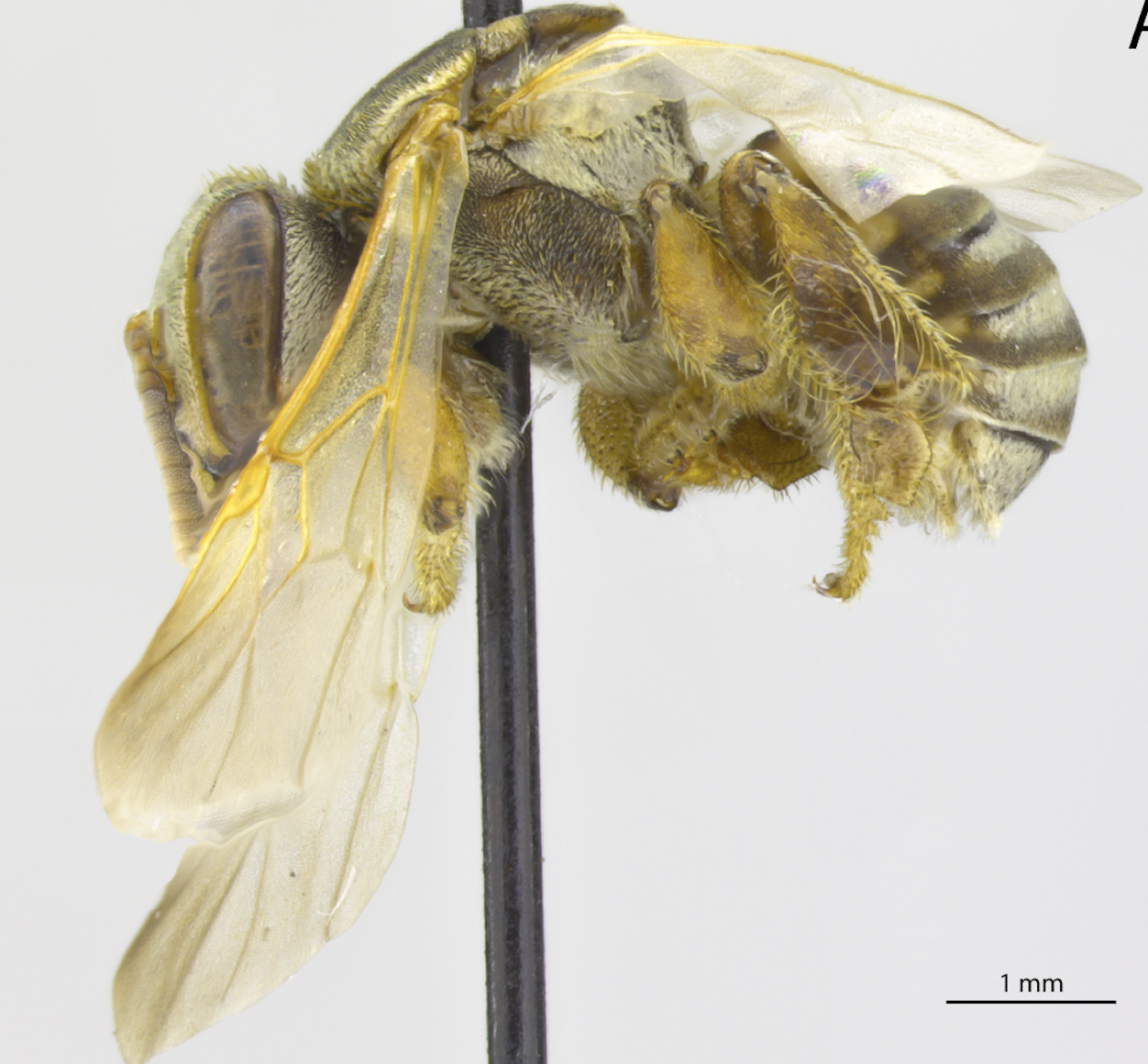}
        \par (g)
    \end{minipage}
    \hfill
    \begin{minipage}[t]{0.3\textwidth}
    \centering
    \includegraphics[width=\linewidth]{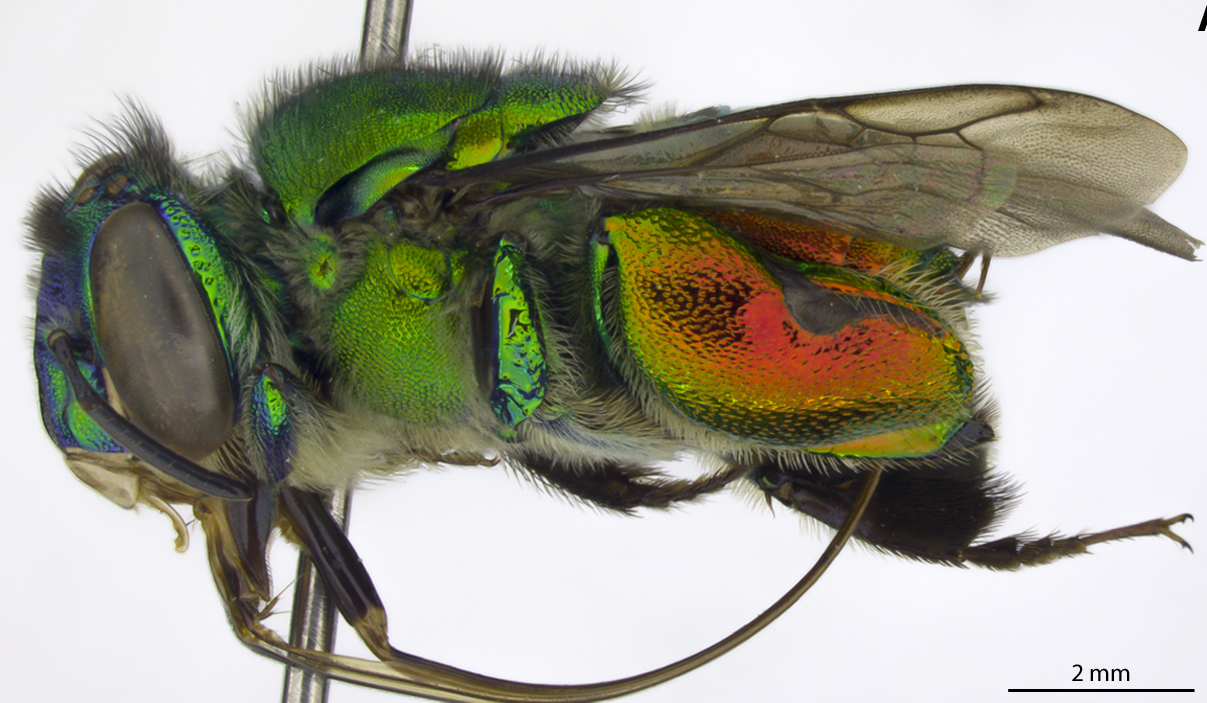}
    \par (h)
    \end{minipage}
    \hfill
    \begin{minipage}[t]{0.3\textwidth}
        \centering
        \includegraphics[width=\linewidth]{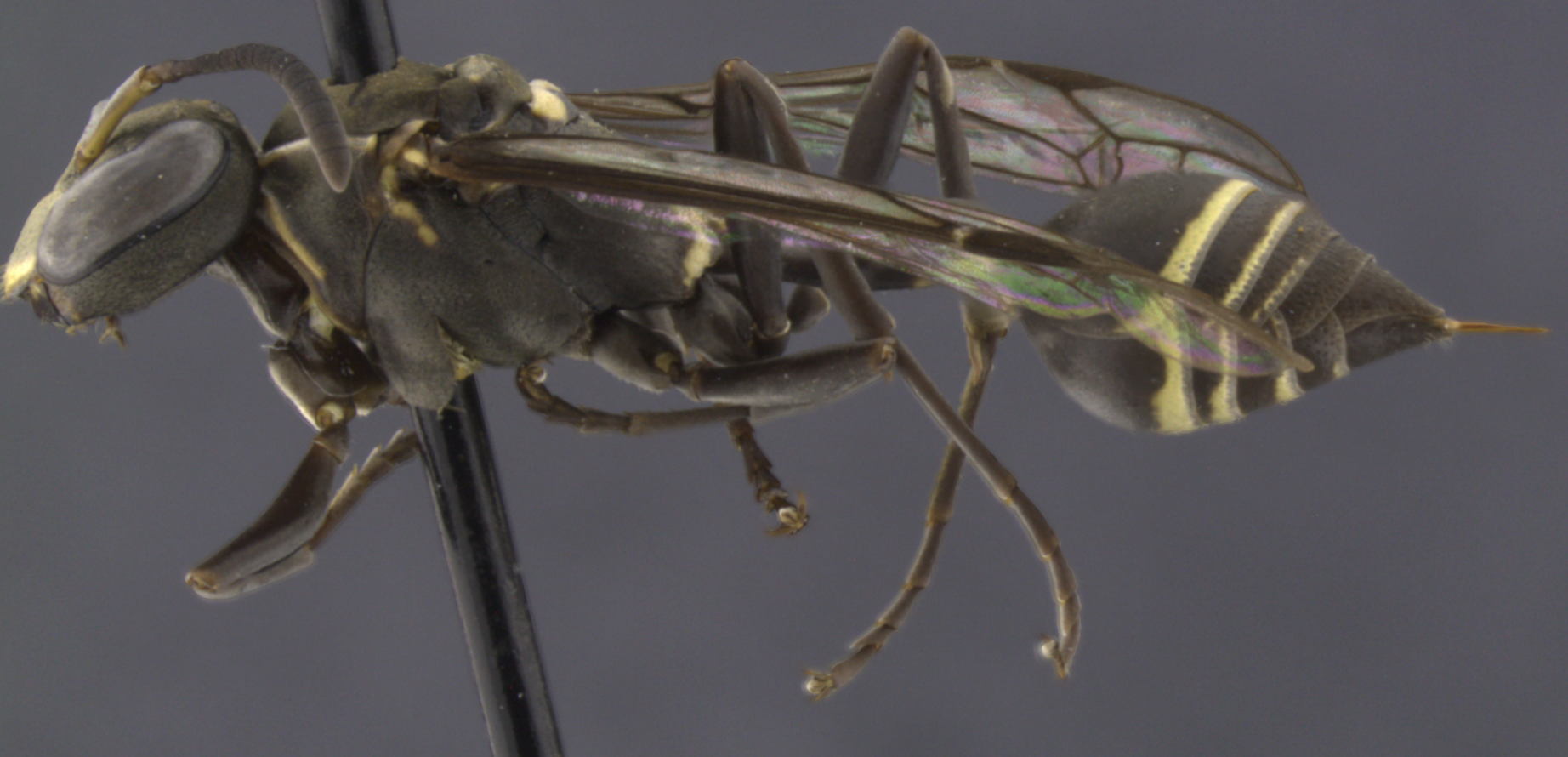}
        \par (i)
    \end{minipage}

    \vspace{0.5cm} 
    
    \begin{minipage}[t]{0.3\textwidth}
        \centering
        \includegraphics[width=\linewidth]{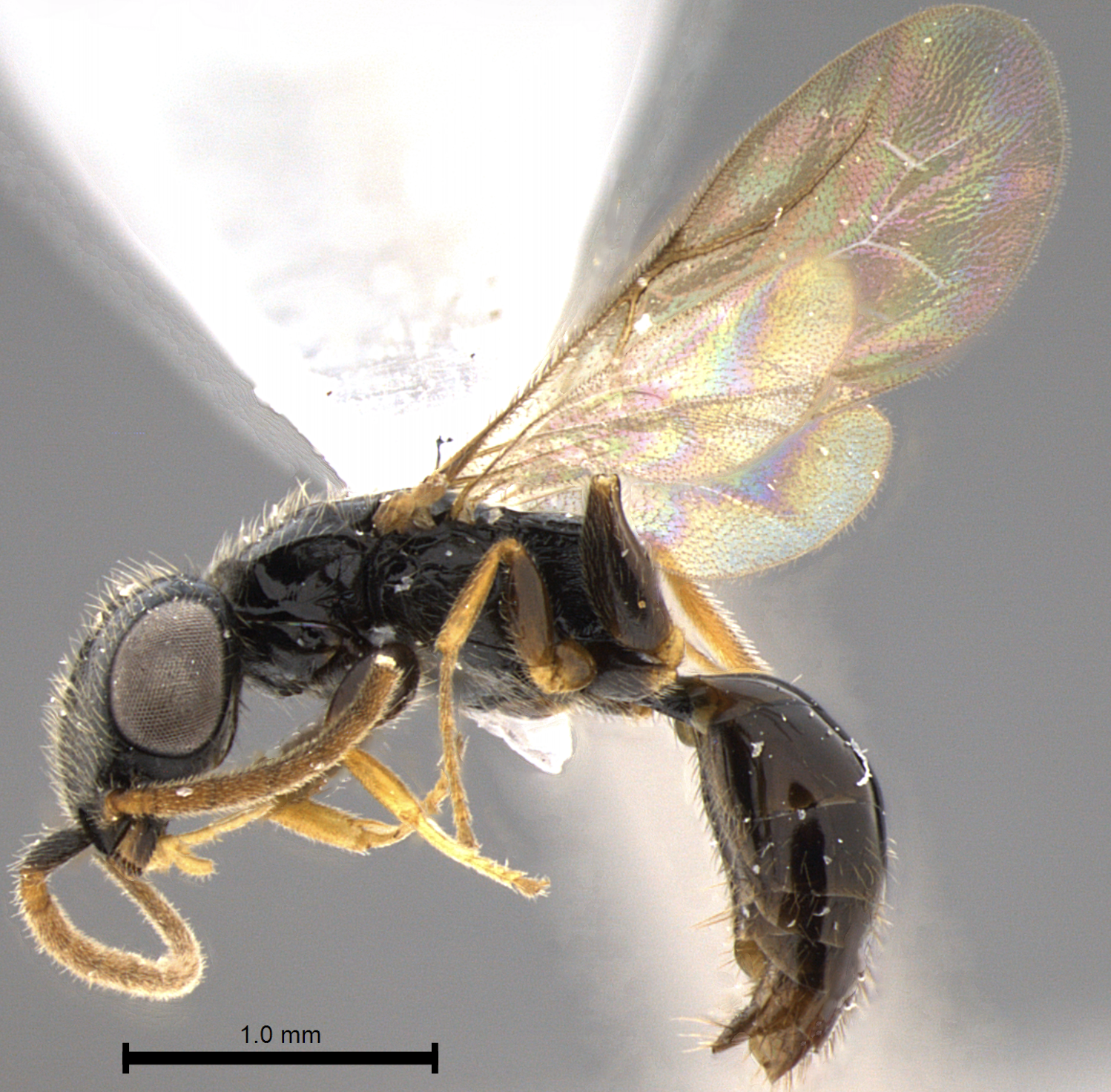}
        \par (j)
    \end{minipage}
    \begin{minipage}[t]{0.35\textwidth}
        \centering
        \includegraphics[width=\linewidth]{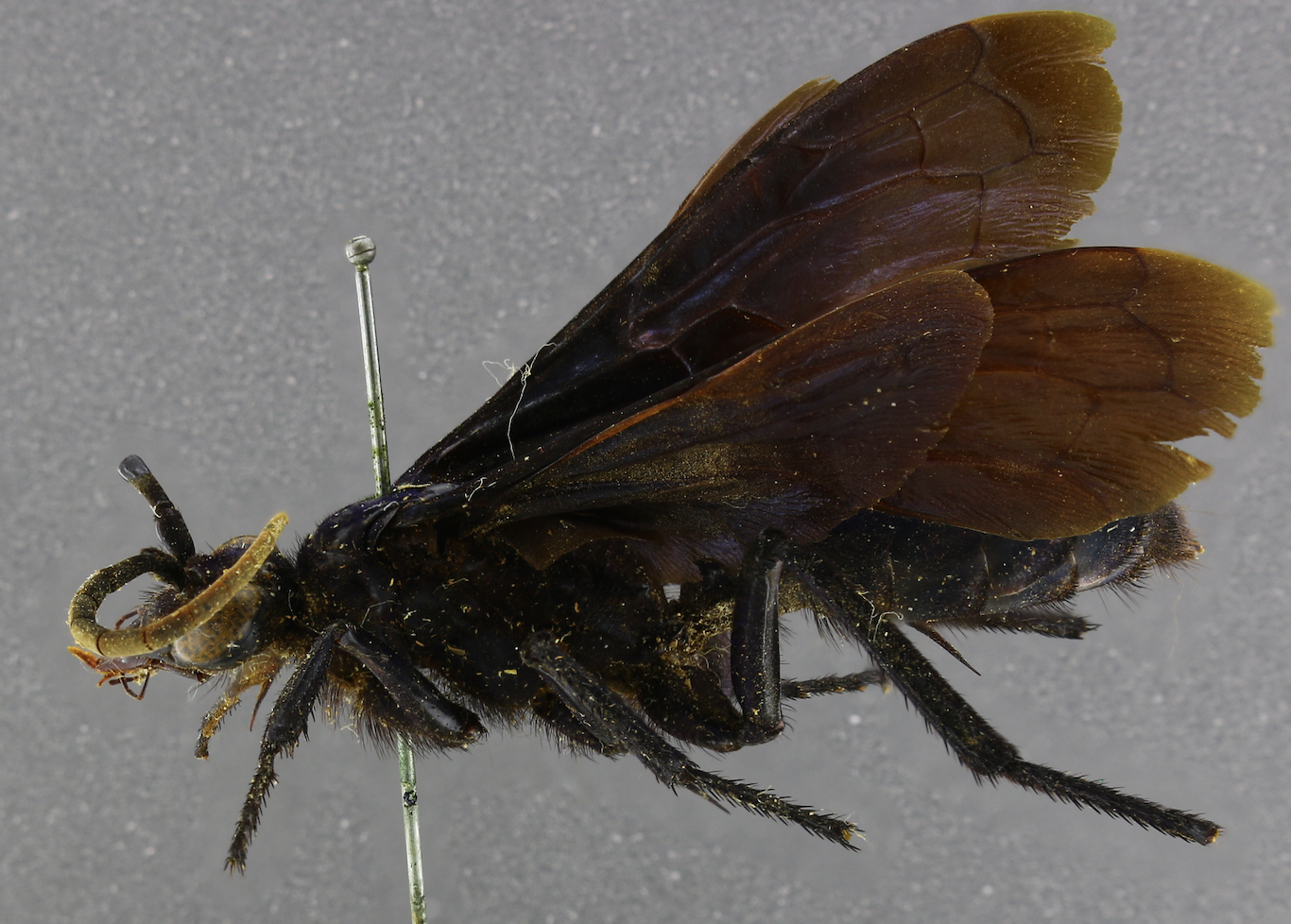}
        \par (k)
    \end{minipage}
    \hfill
    \begin{minipage}[t]{0.3\textwidth}
        \centering
        \includegraphics[width=\linewidth]{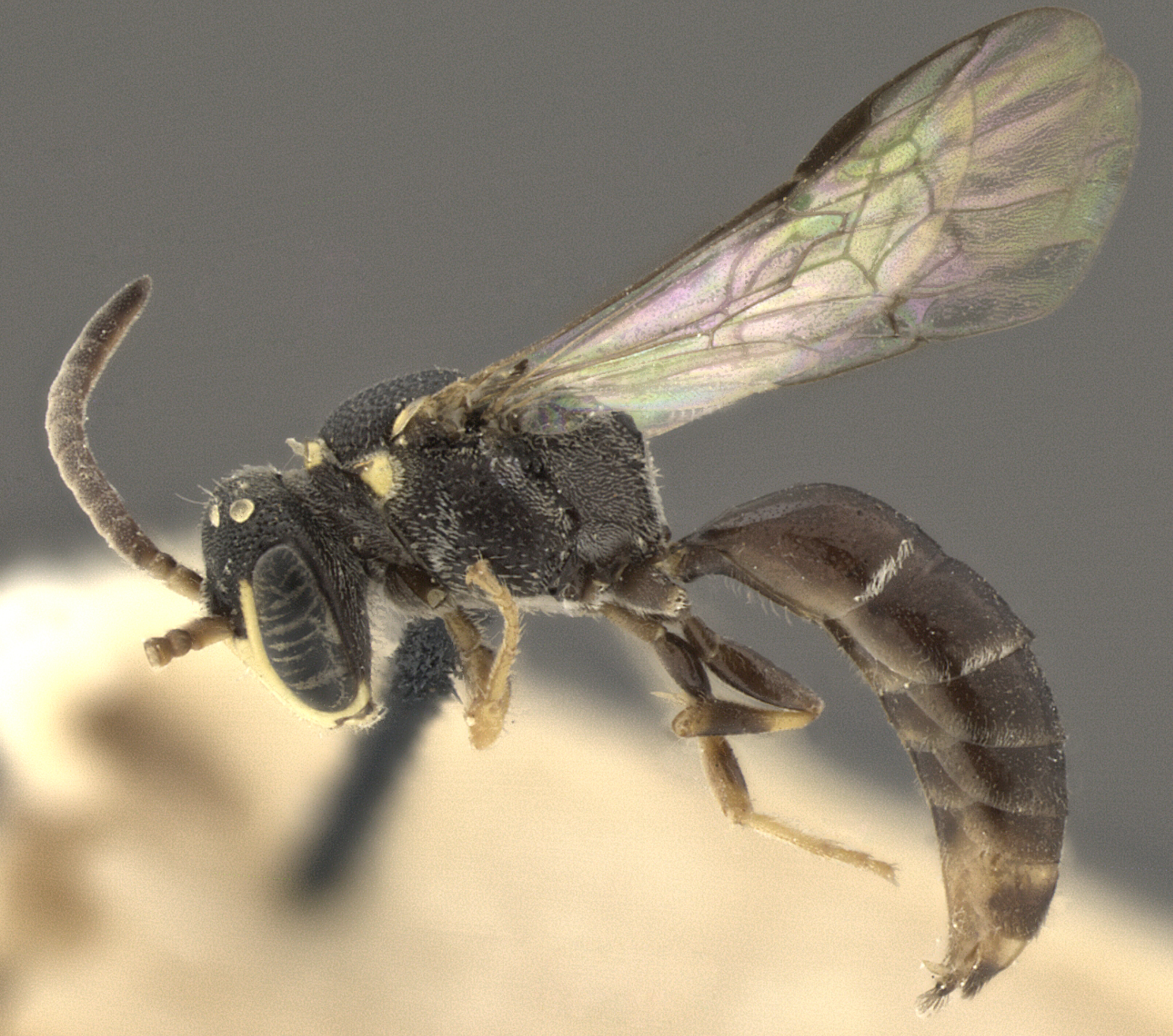}
        \par (m)
    \end{minipage}

    \caption{Examples of morphological variation in the DAPWH dataset. (a), (b), (c) Braconidae; (d), (e), (f) Ichneumonidae; (g), (h) Apidae; (i) Vespidae; (j) Bethylidae; (k) Pompilidae; (m) Colletidae.}
    \label{dataset_example}
\end{figure*}

\begin{figure*}[htp]
    \centering
    \begin{minipage}[t]{0.3\textwidth}
        \centering
        \includegraphics[width=\linewidth]{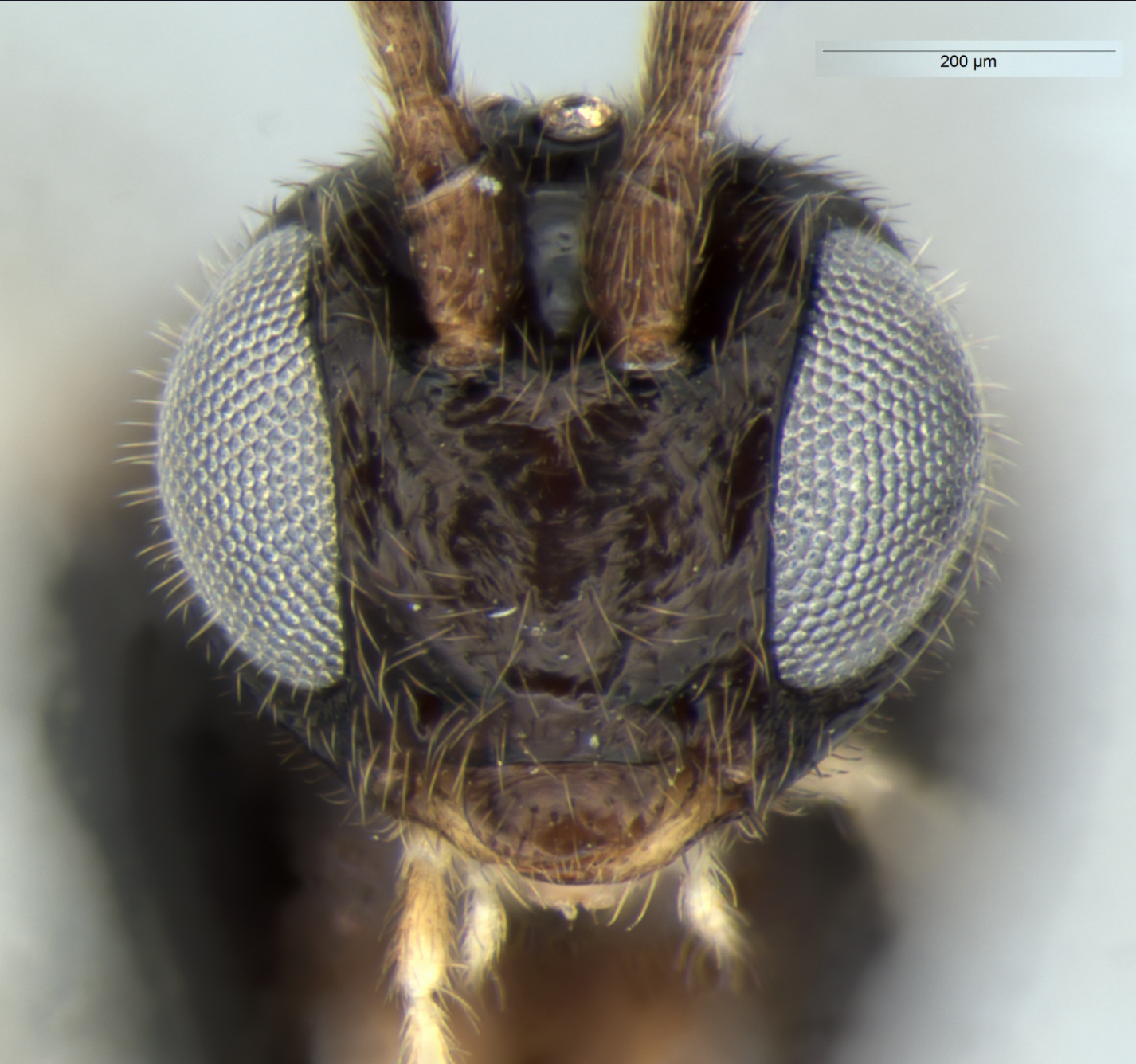}
        \par (a)
    \end{minipage}
    \hfill
    \begin{minipage}[t]{0.35\textwidth}
        \centering
        \includegraphics[width=\linewidth]{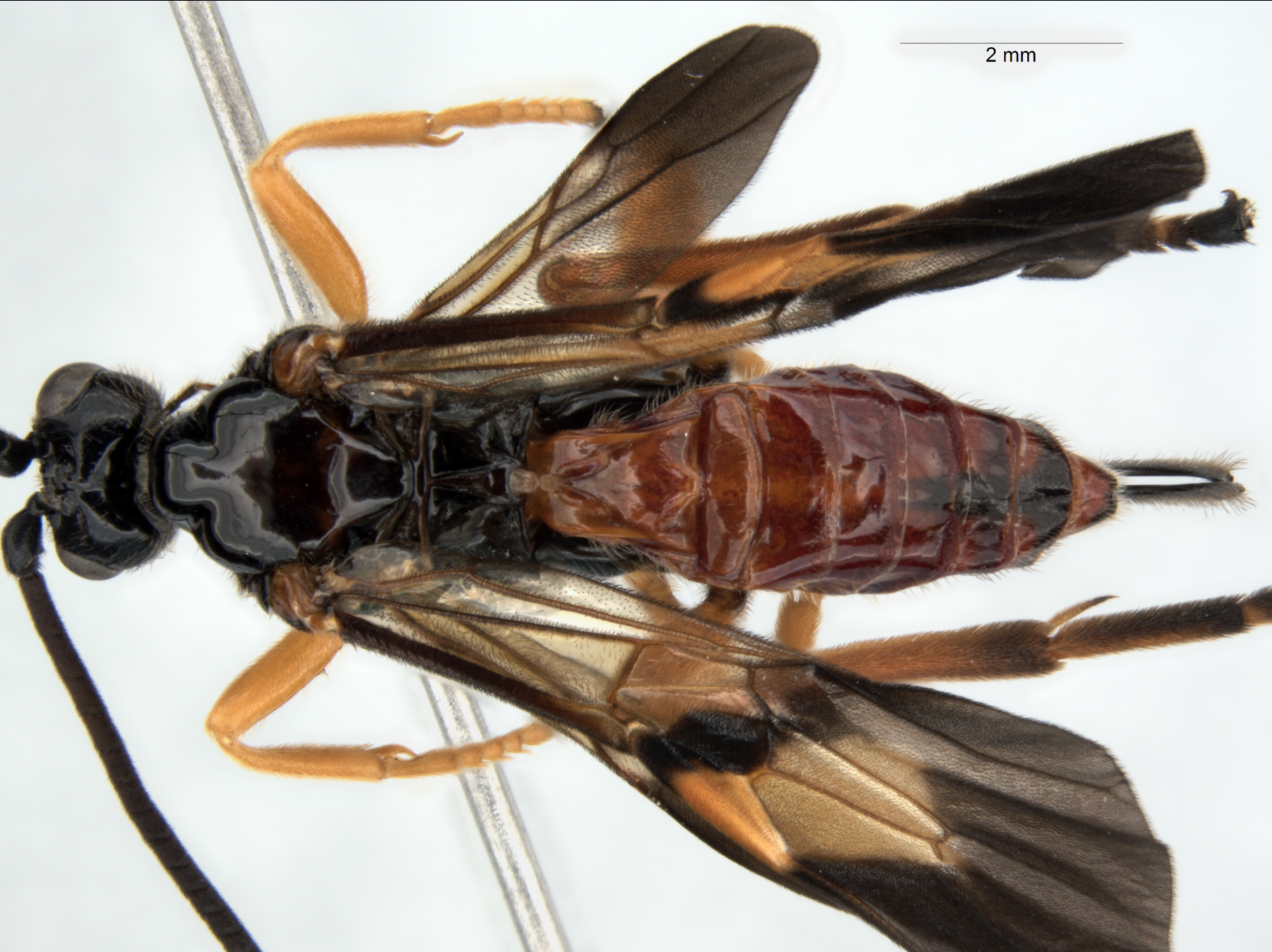}
        \par (b)
    \end{minipage}
    \hfill
        \begin{minipage}[t]{0.3\textwidth}
        \centering
        \includegraphics[width=\linewidth]{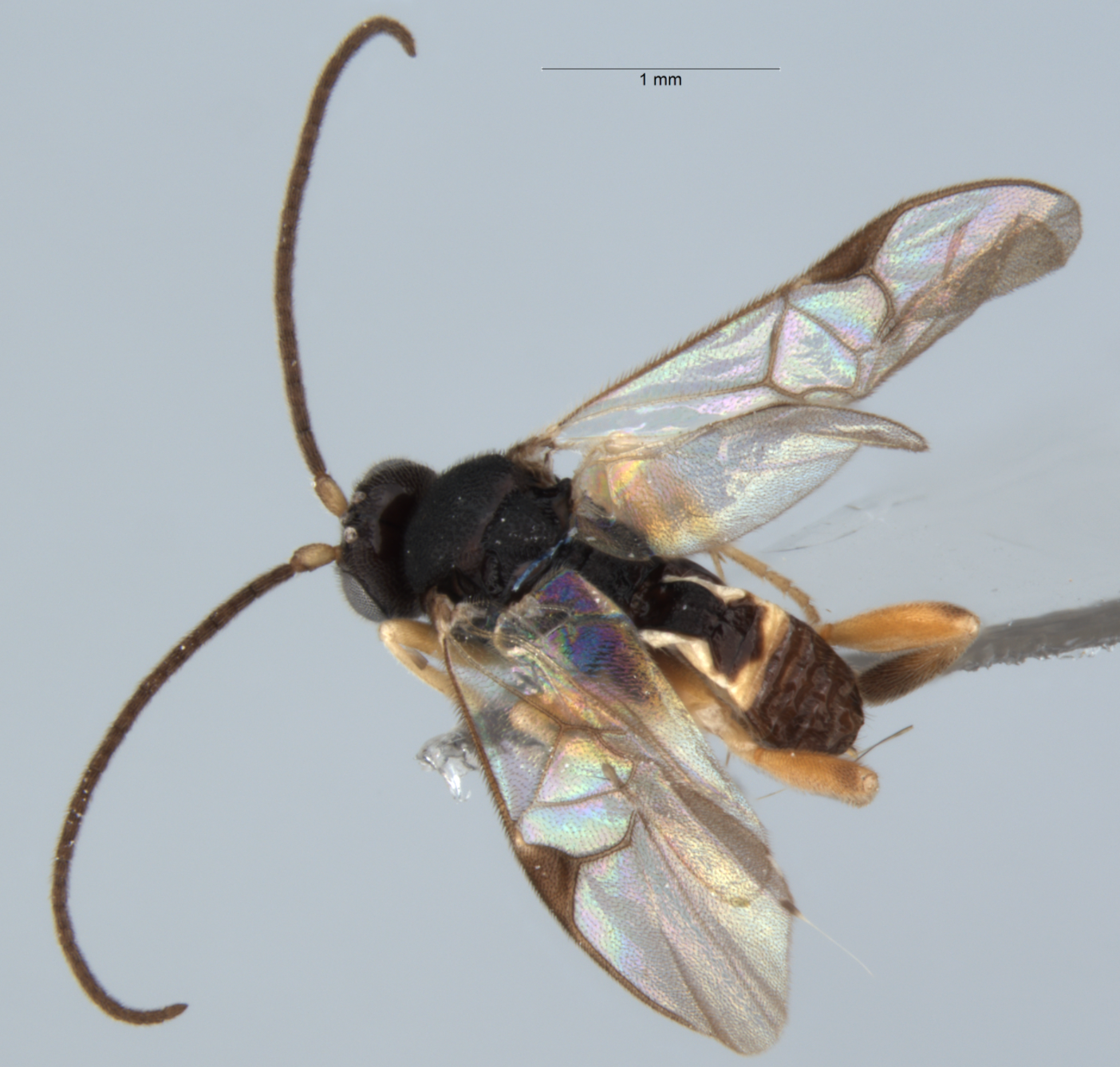}
        \par (c)
    \end{minipage}
    \vspace{0.5cm} 

    \begin{minipage}[t]{0.3\textwidth}
        \centering
        \includegraphics[width=\linewidth]{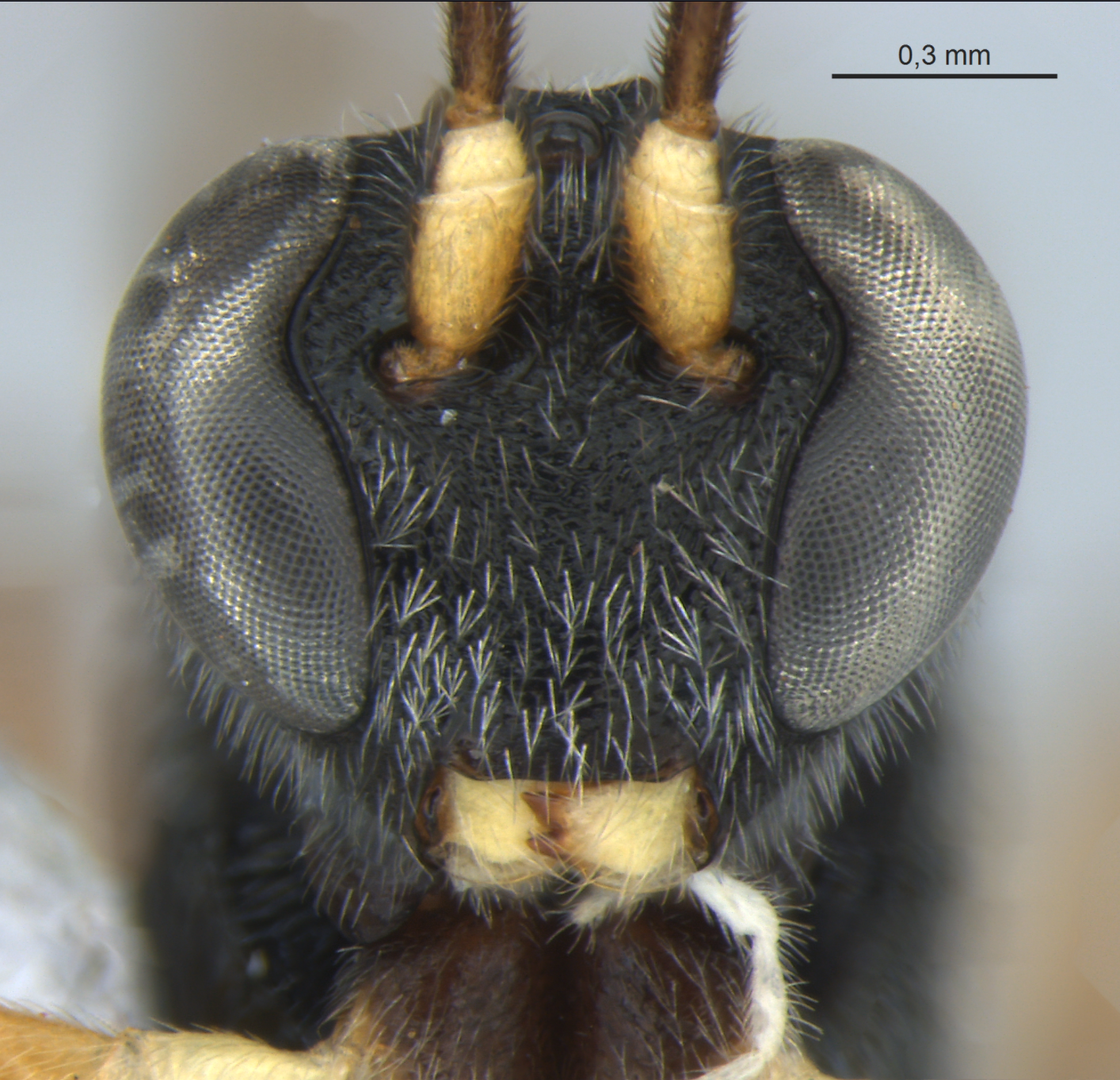}
        \par (d)
    \end{minipage}
    \hfill
        \begin{minipage}[t]{0.35\textwidth}
        \centering
        \includegraphics[width=\linewidth]{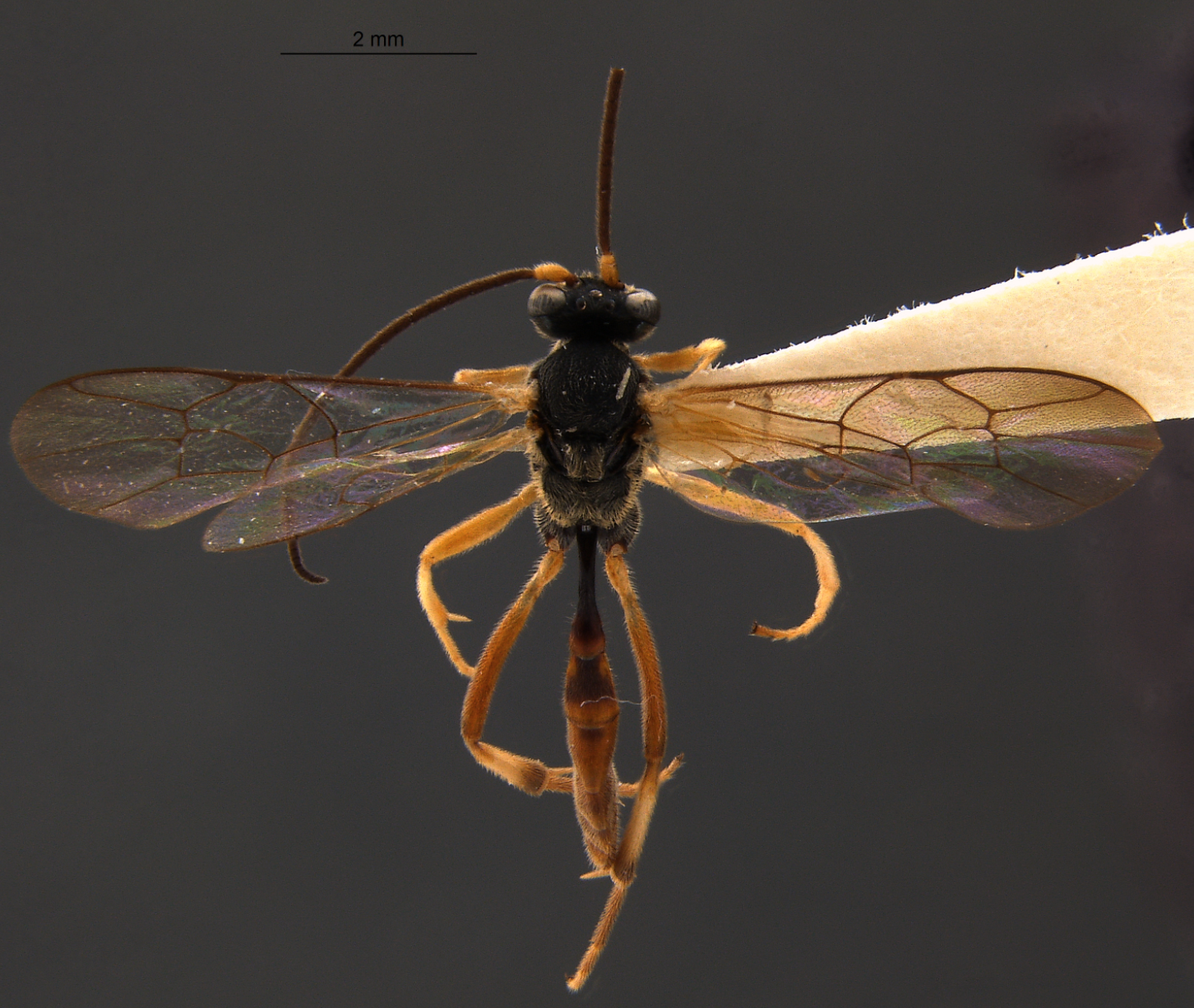}
        \par (e)
    \end{minipage}
    \hfill
        \hfill
        \begin{minipage}[t]{0.3\textwidth}
        \centering
        \includegraphics[width=\linewidth]{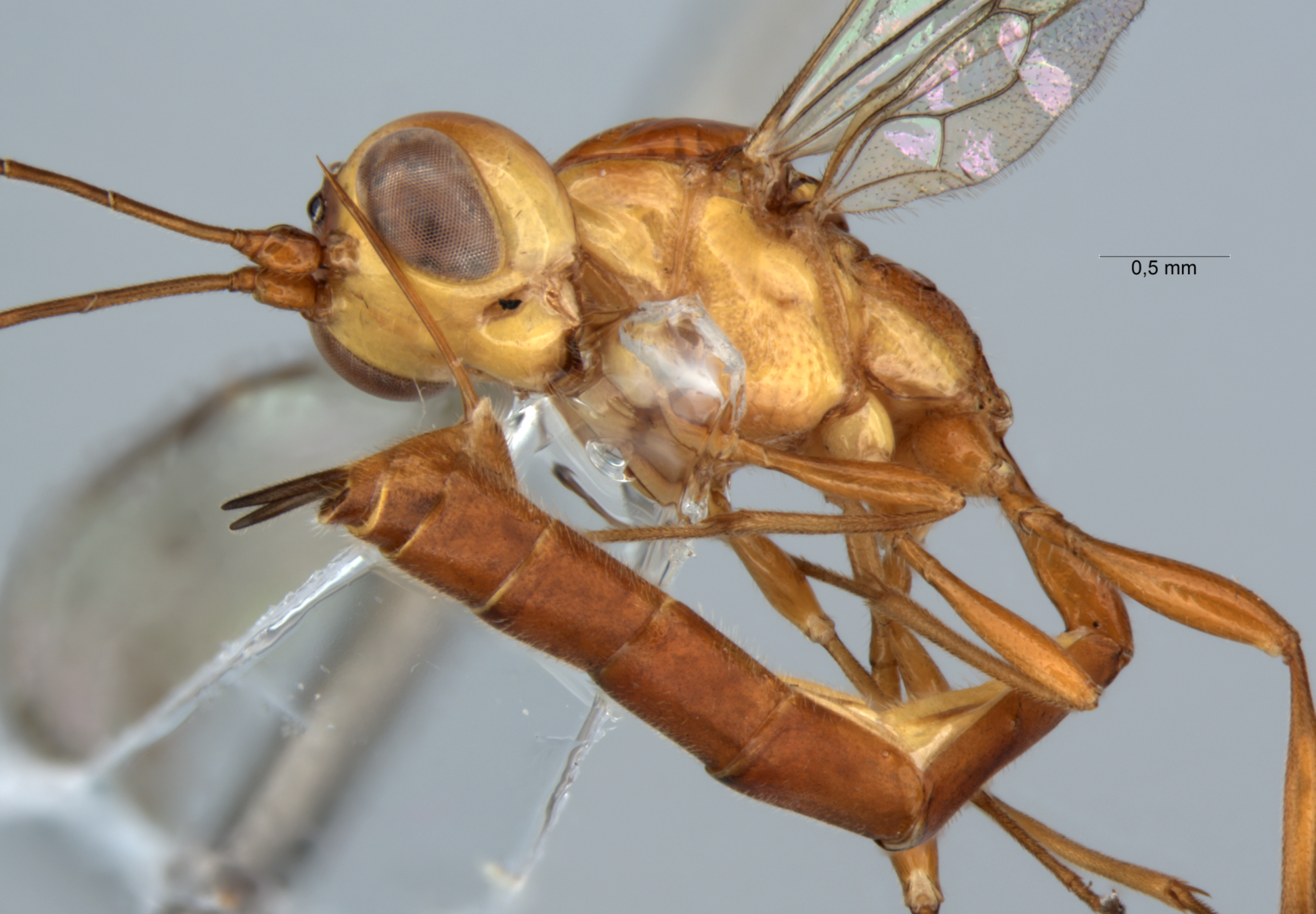}
        \par (f)
    \end{minipage}

    \vspace{0.5cm} 

    \begin{minipage}[t]{0.25\textwidth}
        \centering
        \includegraphics[width=\linewidth]{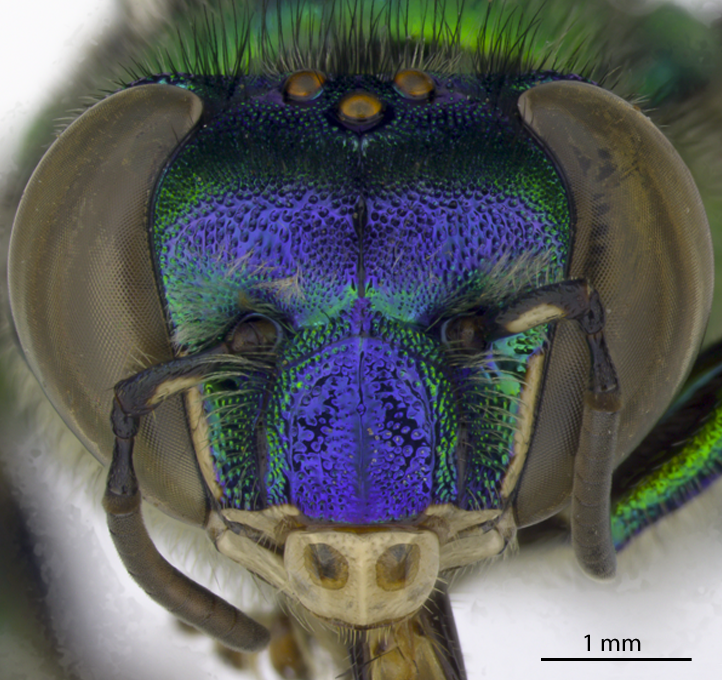}
        \par (g)
    \end{minipage}
    \hfill
    \begin{minipage}[t]{0.35\textwidth}
    \centering
    \includegraphics[width=\linewidth]{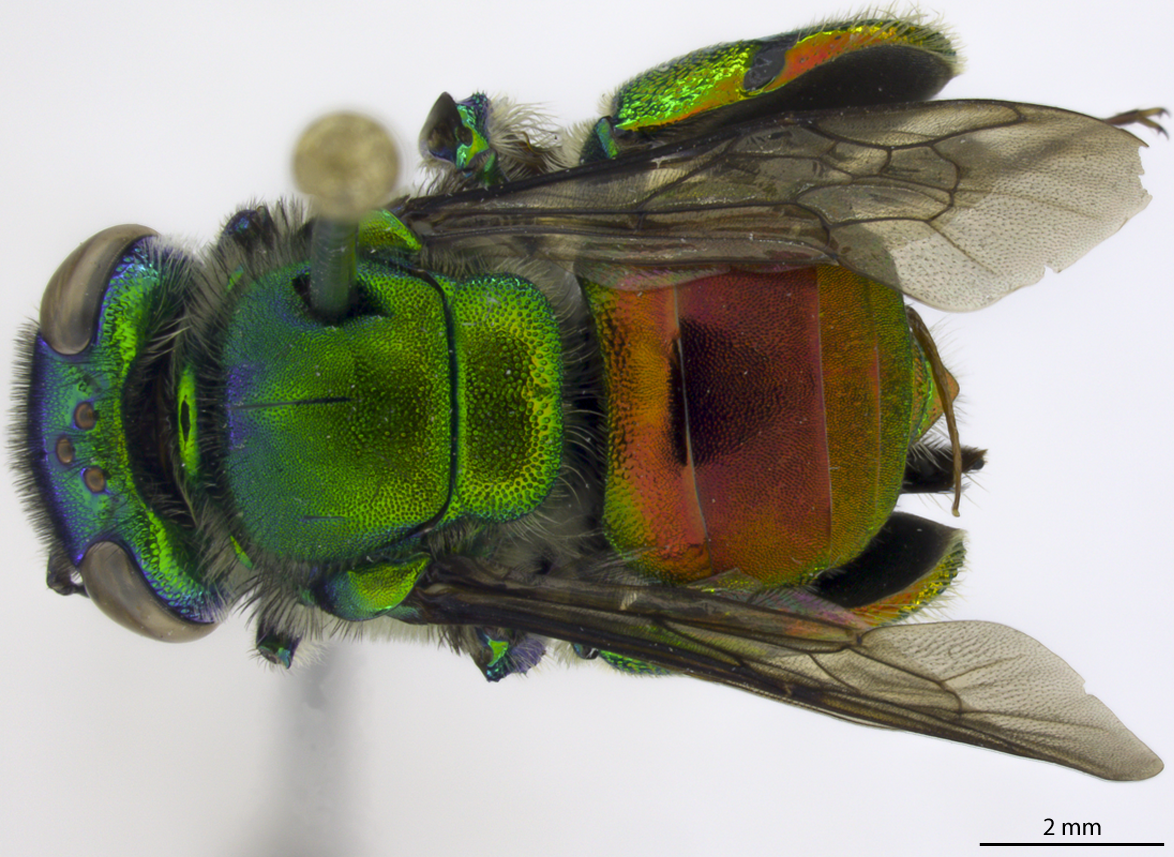}
    \par (h)
    \end{minipage}
    \hfill
    \begin{minipage}[t]{0.3\textwidth}
        \centering
        \includegraphics[width=\linewidth]{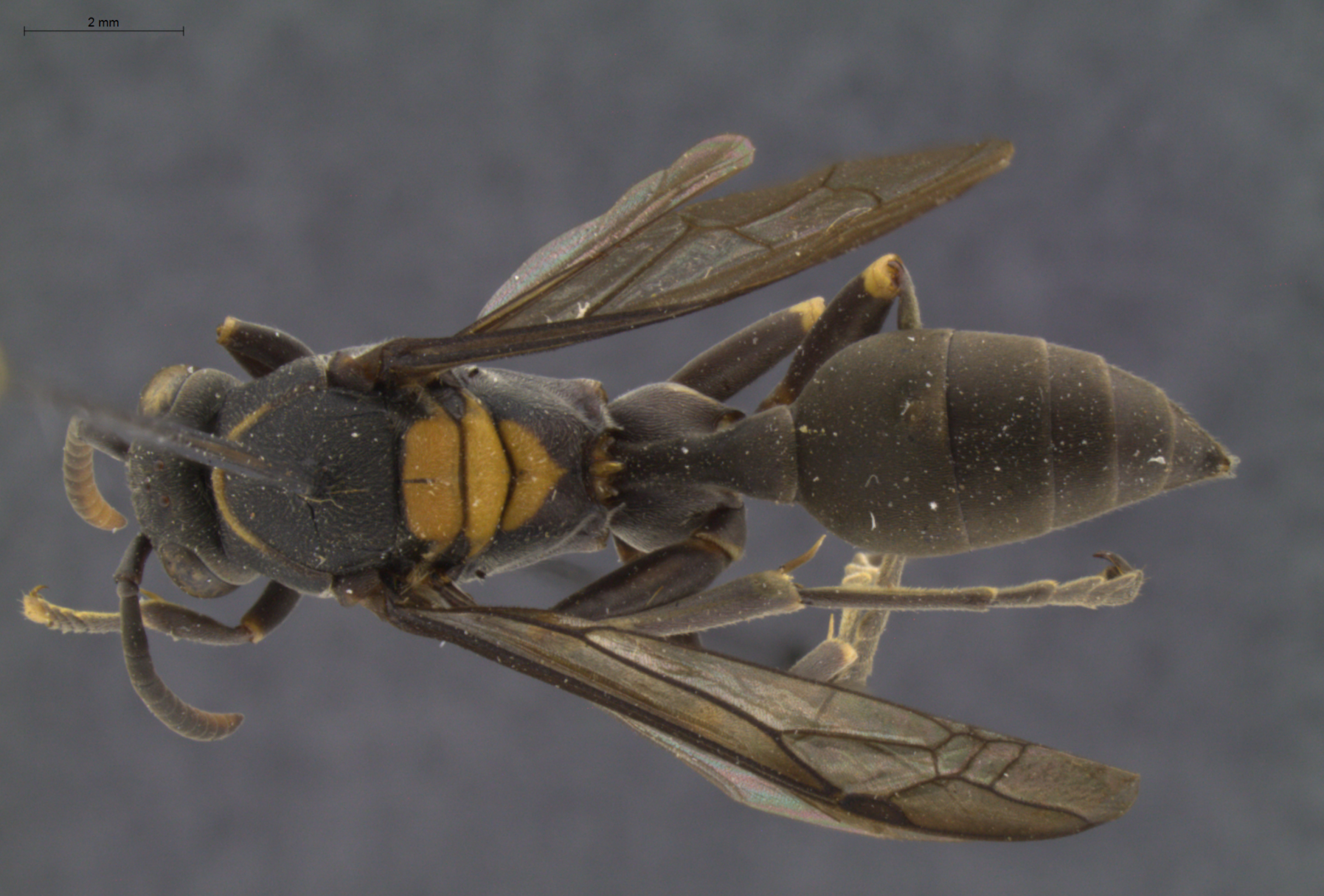}
        \par (i)
    \end{minipage}
    \caption{Examples of morphological variation in the DAPWH dataset. (a), (b), (c) Braconidae; (d), (e), (f) Ichneumonidae; (g), (h) Apidae; (i) Vespidae.}
    \label{dataset_example2}
\end{figure*}

The Apoidea families (Andrenidae, Apidae, Colletidae, Halictidae, Megachilidae) and Chrysididae were primarily sourced from the Coleção Entomológica Prof. J.M.F. Camargo (RPSP) \cite{Almeida_rpsp_2020} provided by Prof. Dr. Eduardo A.B. Almeida, in addition to contributions from 
Museu de Zoologia da Universidade de São Paulo (MZUSP) and the Spencer Collection\cite{needham2025ubcz}. Finally, specimens for Bethylidae, Pompilidae, and Vespidae were acquired from MZUSP \cite{Andrade2018MZSP} and the Spencer Entomological Collection \cite{needham2025ubcz}. Figure \ref{dataset_example} and Figure \ref{dataset_example2} shows example images from the dataset.

The taxonomic and photographic distribution of the DAPWH dataset is detailed in Table \ref{tab:image_dataset}. The dataset comprises a total of 3,556 high-resolution images across 11 Hymenoptera families, organized into three primary morphological perspectives: dorsal/ventral, frontal, and lateral. The images have an average resolution of $1892 \times 1478$ pixels, a standard deviation of $1085 \times 858.$, and a maximum resolution of $8050 \times 5372$ pixels.

The family Ichneumonidae represents the largest portion of the dataset with 786 images, followed closely by Braconidae with 648 images. These two families are of particular interest for automated identification of parasitoid wasps. From a viewpoint perspective, the Lateral view is the most prevalent, containing 1,616 images, providing a substantial foundation for profiling lateral morphological traits.

\subsection*{Annotation Process}
To facilitate the training of detection models, a subset of 1,739 images was selected for detailed labeling and processed using the Computer Vision Annotation Tool (CVAT)\cite{boris_cvat_2020} and the Segment Anything Model (SAM)\cite{kirillov2023segment}. The annotation protocol involved the precise delineation of segment masks and bounding boxes for three distinct classes: the full insect body, wings, and scale bar.

\begin{figure*}[htp]
    \centering
    \includegraphics[width=0.8\textwidth]{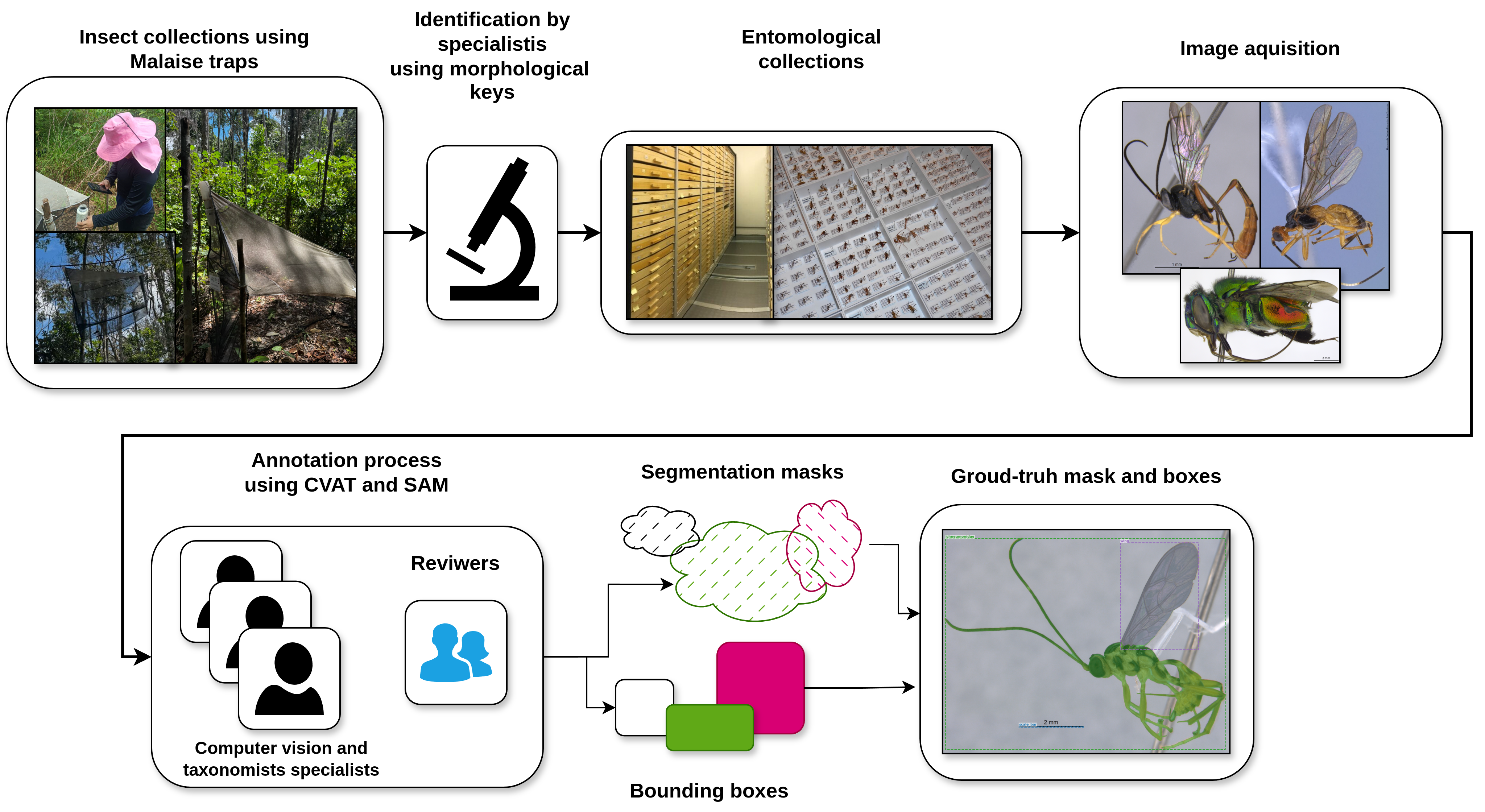}
        \caption{Illustration of the dataset creation and annotation workflow for the subset COCO annotated dataset. Initial segmentation masks and bounding boxes were generated by computer vision experts in CVAT, assisted by the SAM, and subsequently reviewed by taxonomists.}
    \label{fig:flow_images}
\end{figure*}

The initial annotations were performed collaboratively by computer vision specialists and taxonomists, followed by a rigorous final review by a lead taxonomist. Upon validation, the final bounding boxes and segmentation masks were generated, and the dataset was exported in the standard format Common Objects in Context (COCO) \cite{lin2015microsoftcococommonobjects}. The complete pipeline for dataset construction and annotation is depicted in Figure \ref{fig:flow_images}. 

The structural complexity of the dataset is further detailed by the annotation properties summarized in Table \ref{cap5:tab:dataset_composition}. Beyond image-level taxonomic labels, the dataset contains 1,794 high-precision segmentation masks for wings and 1,700 bounding boxes for scale bars. Within the COCO subset, Ichneumonidae and Braconidae remain the most densely annotated taxa, with 521 and 406 mask-level annotations, respectively. 


\begin{table}[htp]
\centering
\caption{Distribution of images per family and view.}
\label{table}
\setlength{\tabcolsep}{3pt}
\begin{tabular}{ccccc}
\hline
        \textbf{Family} & \textbf{Dorsal/Ventral} & \textbf{Frontal} & \textbf{Lateral} & \textbf{Total} \\ 
        \hline
        Ichneumonidae   & 205 & 128 & 453 & \textbf{786} \\
        Braconidae      & 175 & 66  & 407 & \textbf{648} \\
        Apidae          & 153 & 155 & 158 & \textbf{466} \\
        Vespidae        & 230 & 87  & 143 & \textbf{460} \\
        Megachilidae    & 183 & 19  & 96  & \textbf{298} \\
        Chrysididae     & 137 & 31  & 76  & \textbf{244} \\
        Andrenidae      & 154 & 11  & 79  & \textbf{244} \\
        Pompilidae      & 57  & 48  & 85  & \textbf{190} \\
        Bethylidae      & 22  & 0   & 72  & \textbf{94}  \\
        Halictidae      & 26  & 18  & 31  & \textbf{75}  \\
        Colletidae      & 26  & 9   & 16  & \textbf{51}  \\
        \textbf{Total} & \textbf{1,368} & \textbf{572} & \textbf{1,616} & \textbf{3,556} \\ 
\hline
    \end{tabular}
\label{tab:image_dataset}
\end{table}

\begin{table}[htp]
    \centering
    \caption{Summary of COCO dataset and Image Properties}
    \label{cap5:tab:dataset_composition}
    \renewcommand{\arraystretch}{1.2}
    \begin{tabular}{lcl}
        \toprule
        \textbf{Parameter} & \textbf{Value} & \textbf{Notes} \\
\hline
        \multicolumn{3}{l}{\textit{\textbf{Taxonomic Classes}}} \\
        Ichneumonidae   & 521 & Family level (Masks) \\
        Braconidae      & 406 & Family level (Masks) \\
        Apidae          & 158 & Family level (Masks) \\
        Vespidae        & 142 & Family level (Masks) \\
        Megachilidae    & 115 & Family level (Masks) \\
        Andrenidae      & 90  & Family level (Masks) \\
        Pompilidae      & 85  & Family level (Masks) \\
        Chrysididae     & 76  & Family level (Masks) \\
        Bethylidae      & 72  & Family level (Masks) \\
        Halictidae      & 49  & Family level (Masks) \\
        Colletidae      & 25  & Family level (Masks) \\
\hline
        \multicolumn{3}{l}{\textit{\textbf{Annotated Features}}} \\
        Wing            & 1,794 & Segmentation Masks \\
        Scale bar       & 1,700 & Bounding boxes \\
\hline
    \end{tabular}
\end{table}
\section*{VALIDATION AND QUALITY} 
To establish a baseline for the DAPWH dataset, we conducted comprehensive image-level classification and object detection benchmarks. Following established methodological conventions, we adopted a stratified split strategy to ensure taxonomic representation across all subsets. The dataset was partitioned into a training set (70\%), with the remaining 30\% equally distributed between validation and testing sets (15\% each). While no single splitting ratio is universally prescriptive, this configuration was selected to balance model generalization against the bias–variance trade-off, considering the high-resolution nature of the imagery and computational constraints \cite{raschka2020modelevaluationmodelselection}.

All computational experiments were executed using the PyTorch framework on a high-performance workstation equipped with an NVIDIA GeForce RTX 4090 GPU with 24 GB. To preserve the fine-grained morphological details inherent in the macro-photography dataset, a standardized input resolution of $224 \times 224$ and $1280 \times 1280$ pixels was maintained across the classification and detection models, respectively.

\subsection*{Image-Level Identification}
To evaluate the discriminative power of different feature extractors on the Hymenoptera dataset, we conducted a comparative analysis of six distinct architectures: EfficientNetV2 \cite{tan2021efficientnetv2smallermodelsfaster}, ConvNeXt \cite{liu2022convnet2020s}, ViT-B/16 \cite{dosovitskiy2021imageworth16x16words}, VGG16\cite{simonyan2015deepconvolutionalnetworkslargescale}, ResNet50 \cite{7780459}, and YOLOv12 \cite{tian2025yolov12attentioncentricrealtimeobject}.

\begin{figure}[htp]
    \centering
    \includegraphics[width=0.53\textwidth]{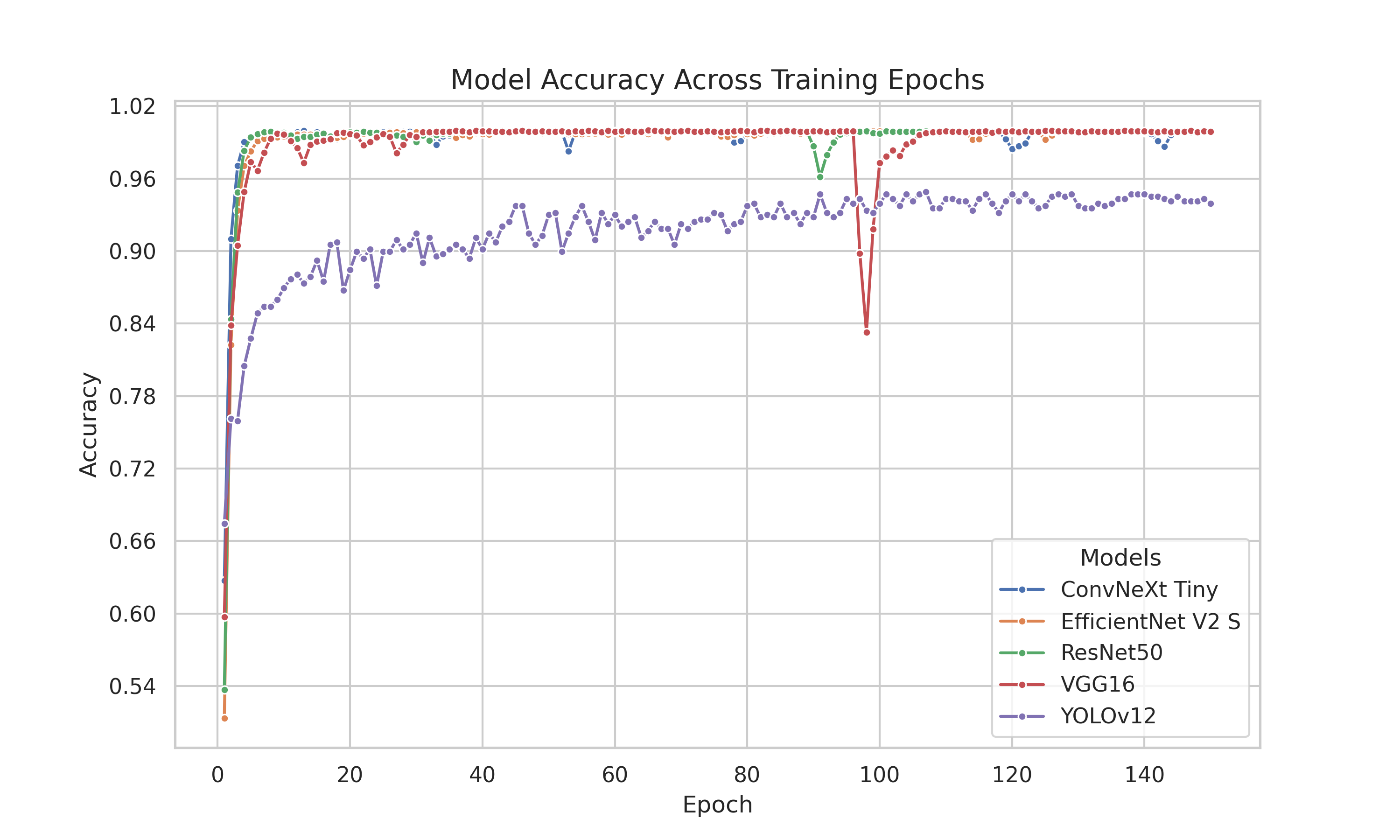} 
    \caption{Validation accuracy per epoch across 150 epochs.}
    \label{fig:learning_curve_class_def}
\end{figure}

The quantitative performance metrics for the image-level classification task are consolidated in Table \ref{v1_cls_accuracy}. The evaluated architectures achieved high efficacy, with Top-1 test-set accuracy ranging from 89.34\% to 92.28\%. Specifically, EfficientNetV2 and ConvNeXt reached a peak accuracy of 92.28\%, while YOLOv12 demonstrated the highest overall robustness with an F1-score of 95.59\%. Furthermore, the convergence behavior of these models is illustrated in Figure \ref{fig:learning_curve_class_def}, which shows the validation accuracy over 150 epochs of training.

\begin{table}[htp]
\centering
\caption{Comparison of validation metrics for the evaluated classification models on the test dataset.}
\begin{tabular}{ccccc}
\hline
Model & Acc. & F1 & Precision & Recall \\
\hline
EfficientNetV2    & 92.2\% & 84.07\% & 93.69\% & 82.54\% \\
ConvNeXt Tiny     & 92.2\% & 87.00\% & 92.38\% & 86.24\% \\
ViTB16            & 89.3\% & 80.45\% & 90.16\% & 78.48\% \\
VGG16             & 89.5\% & 86.39\% & 88.88\% & 84.60\% \\
ResNet50          & 90.1\% & 84.51\% & 88.71\% & 82.10\% \\
YOLOv12           & 91.7\% & 95.59\% & 85.73\% & 88.88\% \\
\hline
\end{tabular}
\label{v1_cls_accuracy}
\end{table}


\subsection*{Baseline Detector}
The baseline detector model served as the baseline for object detection, utilizing the standard hyperparameters recommended by the YOLOv8 \cite{10533619} and YOLOv12 \cite{tian2025yolov12attentioncentricrealtimeobject} literature to establish a performance benchmark.

\begin{figure*}[htp]
    \centering
    \includegraphics[width=0.85\textwidth]{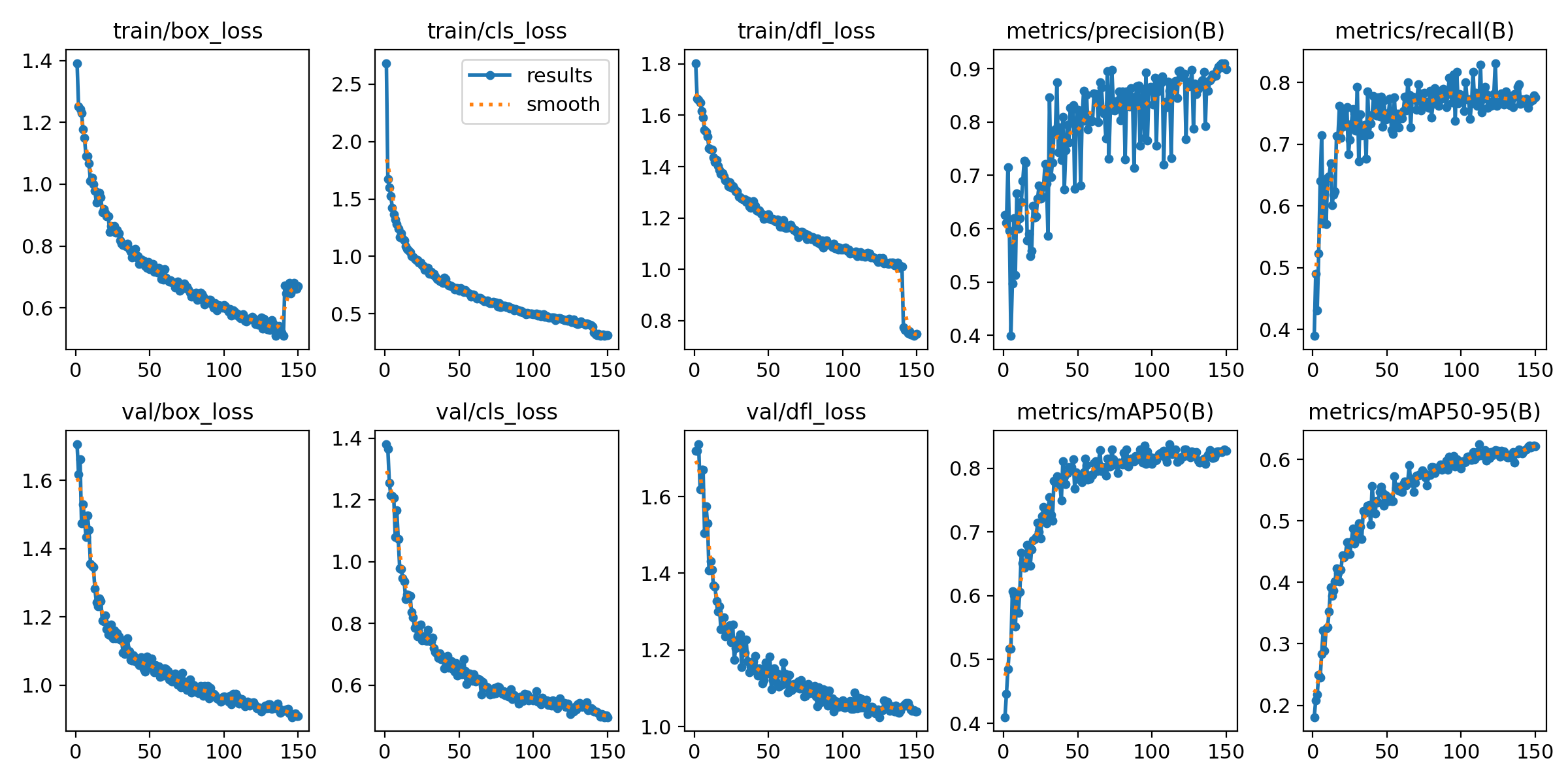} 
    \caption{Training and validation metrics over 150 epochs for the YOLOv8 detection model.}
    \label{detection_curves_yolo8}
\end{figure*}

\begin{figure*}[htp]
    \centering
    \includegraphics[width=0.85\textwidth]{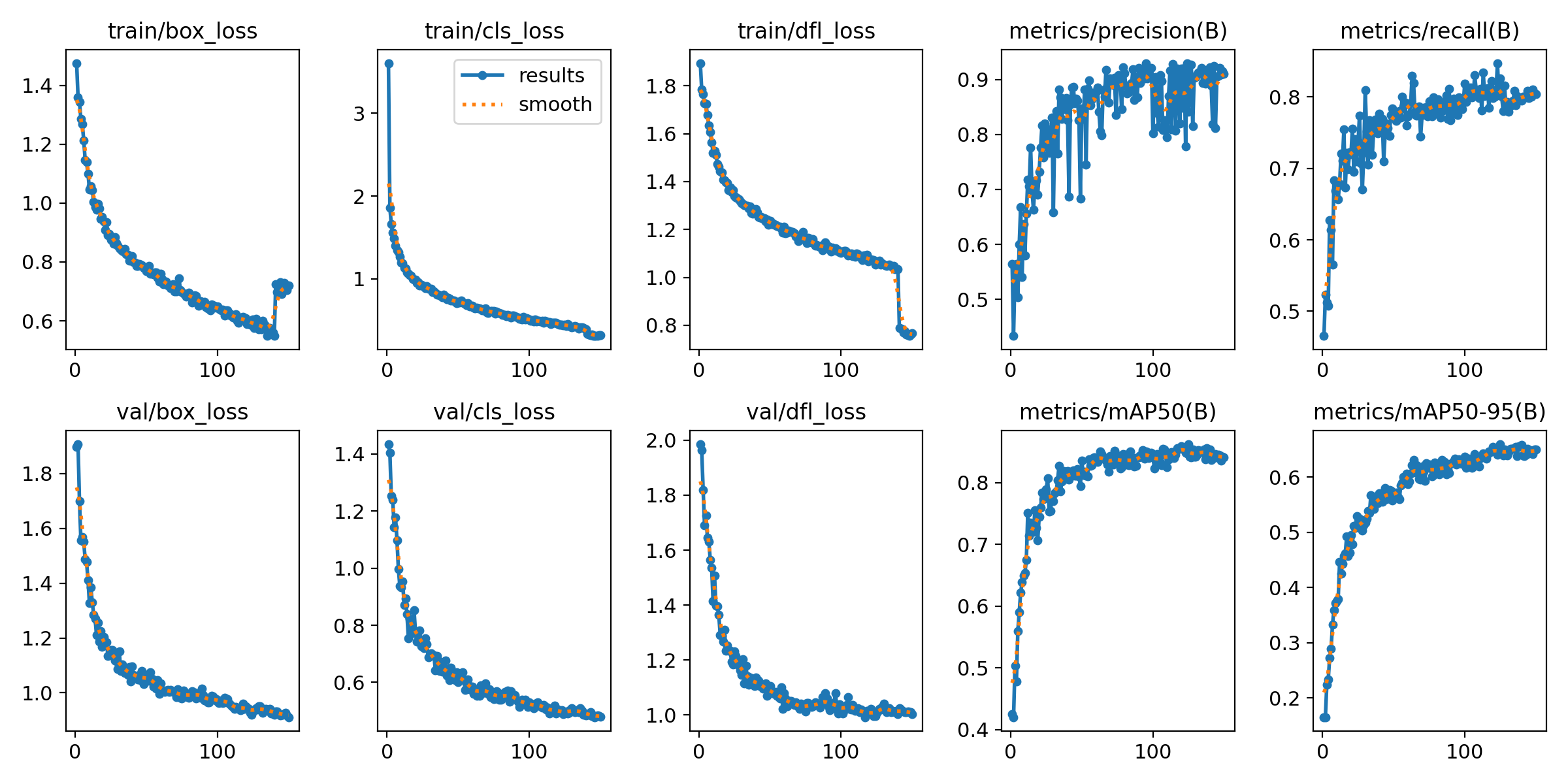}
    \caption{Training and validation metrics over 150 epochs for the YOLOv12 detection model.}
    \label{detection_curves_yolo12}
\end{figure*}

The object detection model was trained for 150 epochs, with performance metrics monitored across the validation set. Figure \ref{detection_curves_yolo8} and Figure \ref{detection_curves_yolo12} illustrate the training dynamics, including loss convergence and metric evolution. Both training and validation losses demonstrated a consistent downward trend, indicating that the model effectively learned to localize bounding boxes and classify insect families without significant overfitting. The final performance metrics for the object detection task, evaluated on the COCO subset test dataset, are summarized in Table \ref{tab:detection_test_results}.

\begin{table}[htp]
\centering
\caption{Comparative performance metrics for object detection architectures on the DAPWH test dataset.}
\label{tab:detection_test_results}
\begin{tabular}{cccc}
\hline
Model & mAP\@50  & Precision & Recall \\
\hline
YOLOv8  & 86.26\%  & 79.08\%  & 85.83\%  \\
YOLOv12 & 90.53\%  & 90.23\%  & 82.54\%  \\
\end{tabular}
\end{table}

\begin{figure*}[htp]
    \centering
    \includegraphics[width=1\textwidth]{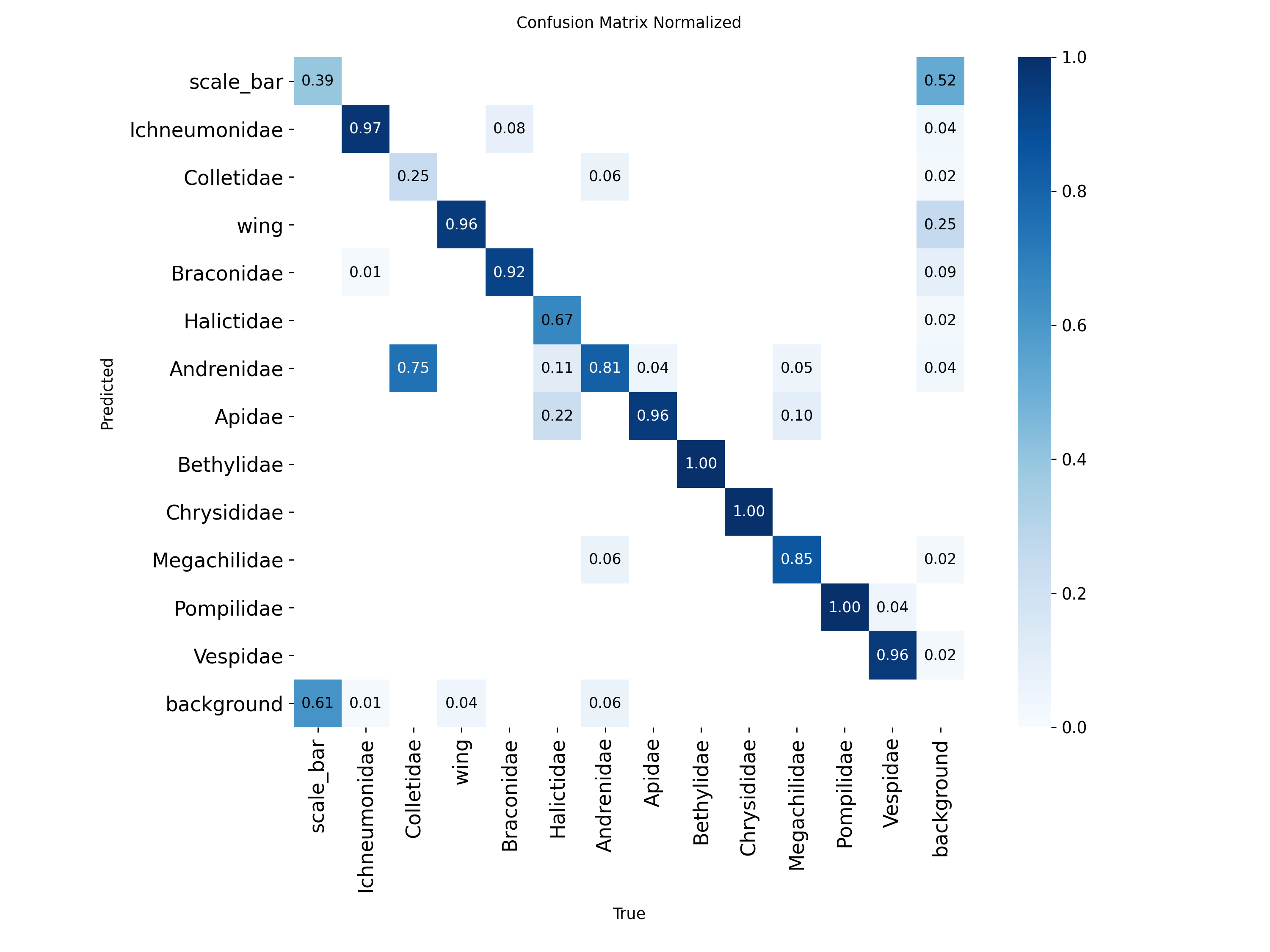} 
    \caption{Confusion matrix normalized for YOLOv12.}
    \label{confusion_matrix_yolov12}
\end{figure*}

The normalized confusion matrix for the YOLOv12s model is illustrated in Figure \ref{confusion_matrix_yolov12}. The diagonal elements represent the recall for each category, demonstrating the model's strong ability to identify key taxonomic families and morphological features. Notably, the model achieved exceptional classification accuracy for Ichneumonidae, Apidae, Vespidae, and Bethylidae.

The YOLOv12 model demonstrated high spatial and taxonomic precision, particularly for the identification of morphological features such as wings, which were detected with an accuracy of 96\%. However, localized classification errors were observed between the families Andrenidae and Colletidae. This discrepancy is largely due to the significant class imbalance in the COCO subset; while the model achieved high recall for well-represented taxa, the Colletidae family contains only 25 mask-level annotations, representing the smallest sample size in the dataset. Such limited taxonomic representation constrains the model's ability to learn the distinct morphological boundaries of this family, especially when compared to the 927 images available for the superfamily Ichneumonoidea. Additionally, the scale\_bar class shows a 0.61 background false negative rate, suggesting that while the object is detected, its localization boundaries occasionally overlap with background noise in high-resolution frames.

\begin{figure*}[htp]
    \begin{minipage}[t]{0.3\textwidth}
        \centering
        \includegraphics[width=\linewidth]{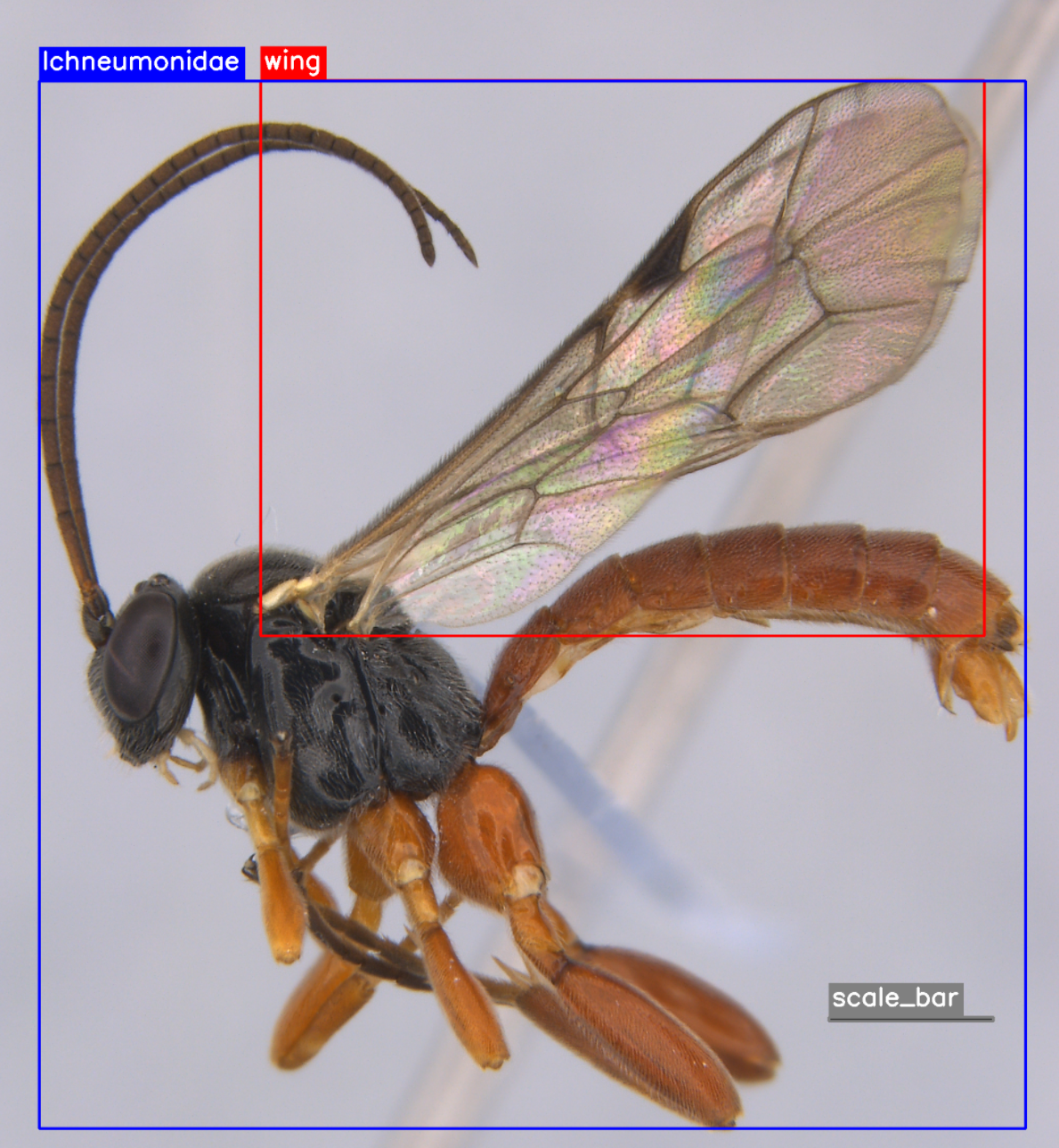}
        \par (a)
    \end{minipage}
    \hfill
    \begin{minipage}[t]{0.3\textwidth}
        \centering
        \includegraphics[width=\linewidth]{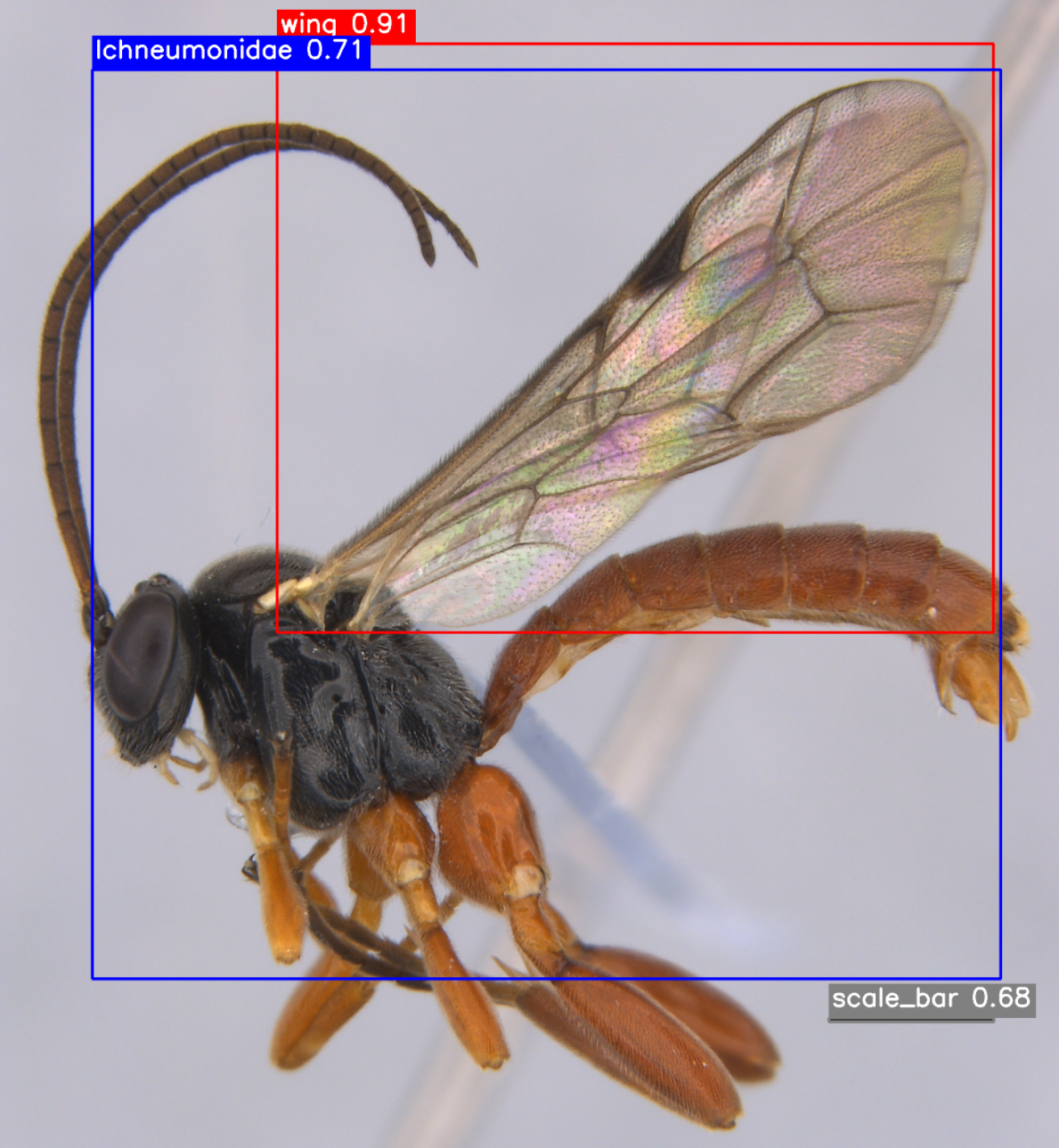}
        \par (b)
    \end{minipage}
    \hfill
    \begin{minipage}[t]{0.3\textwidth}
        \centering
        \includegraphics[width=\linewidth]{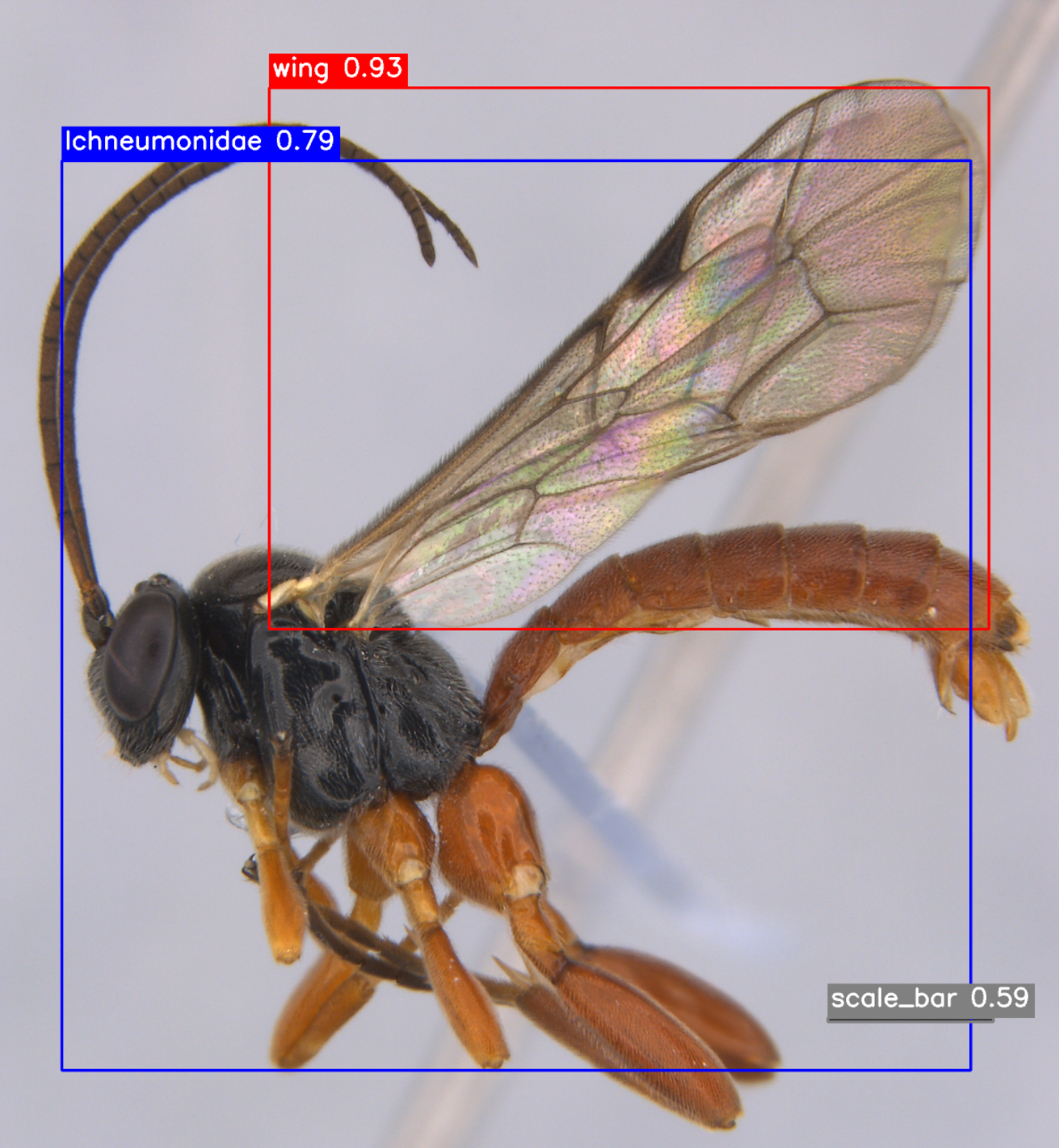}
        \par (c)
    \end{minipage}
    \begin{minipage}[t]{0.3\textwidth}
        \centering
        \includegraphics[width=\linewidth]{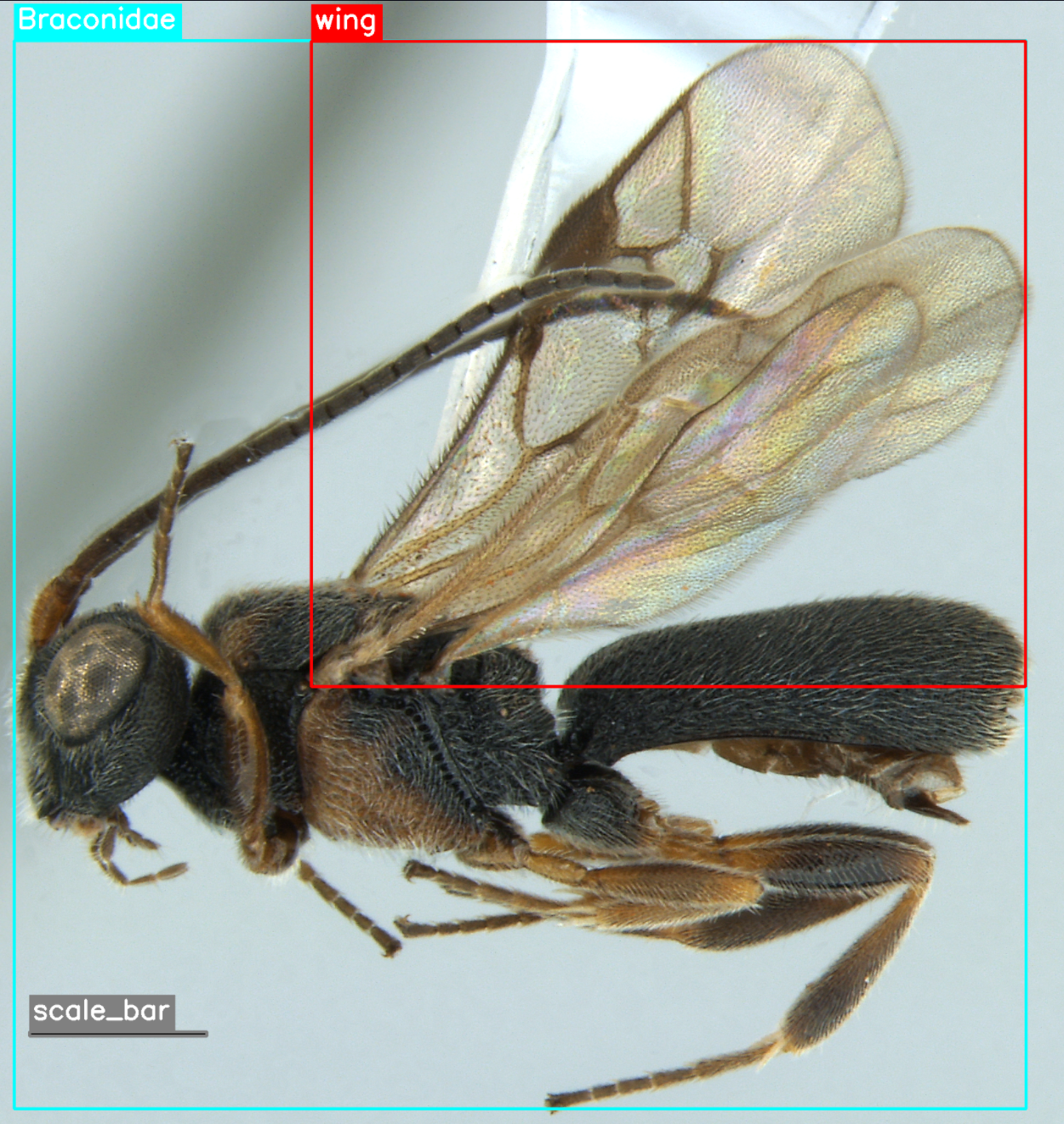}
        \par (d)
    \end{minipage}
    \hfill
    \begin{minipage}[t]{0.3\textwidth}
        \centering
        \includegraphics[width=\linewidth]{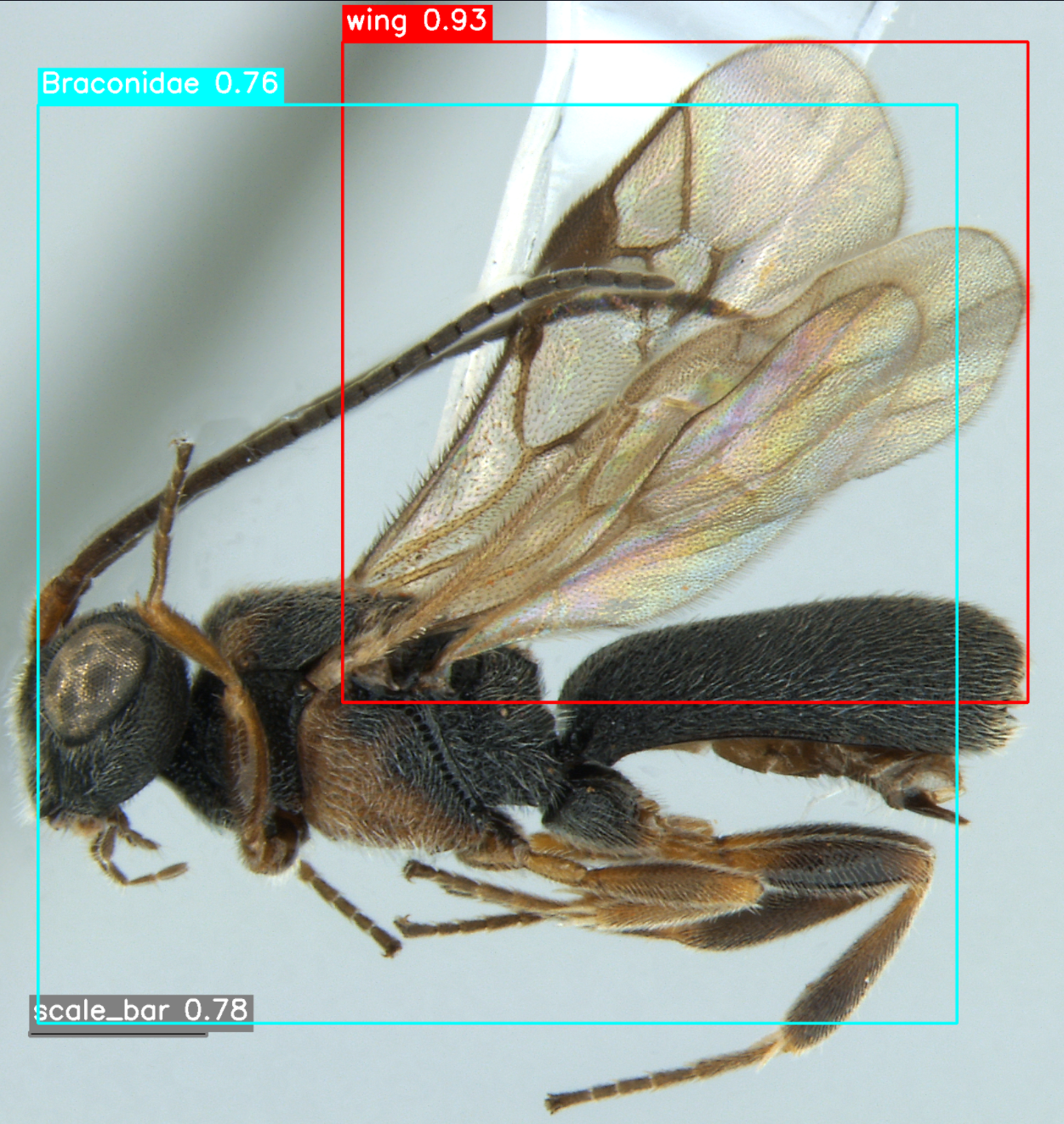}
        \par (e)
    \end{minipage}
    \hfill
    \begin{minipage}[t]{0.3\textwidth}
        \centering
        \includegraphics[width=\linewidth]{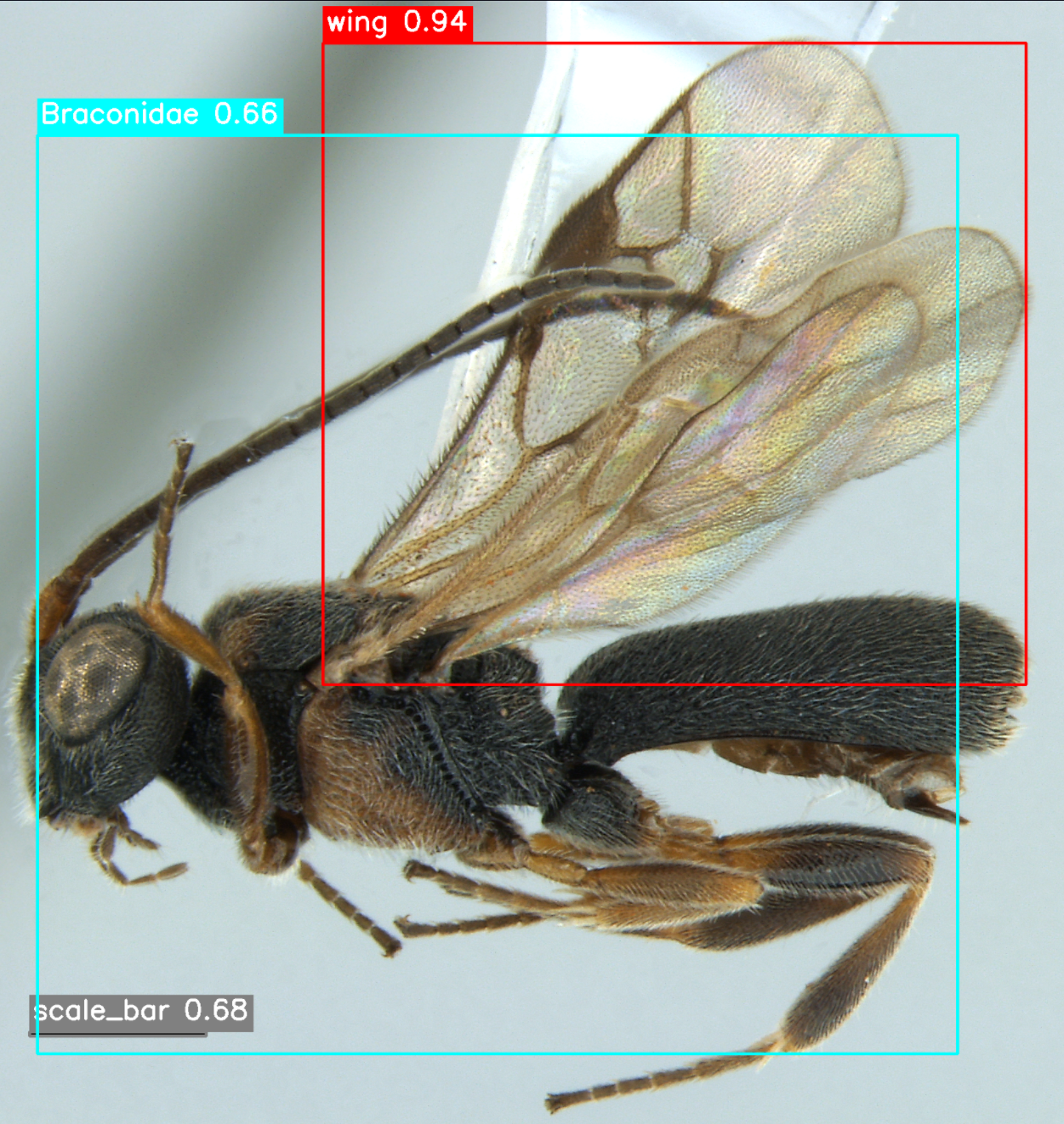}
        \par (f)
    \end{minipage}
    \caption{Visual comparison of detection results. (a,d) The corresponding ground truth annotations; (b,e) The predicted bounding boxes generated by the YOLOv8 during inference; (c,f) The predicted bounding boxes generated by the YOLOv12 during inference.}
    \label{detection_example}
\end{figure*}

Figures \ref{detection_example} present representative inference detection examples from the test dataset. As illustrated, the object detection model shows strong spatial alignment, with the predicted bounding boxes for both the insect and wing structures closely matching the ground truth annotations. 
\section*{RECORDS AND STORAGE} 

The DAPWH \cite{herrera_pinheiro_2026_18501018} dataset, comprising images of Hymnoptera families, is archived and publicly accessible via Zenodo (\hyperlink{https://www.doi.org/10.5281/zenodo.18501018}{10.5281/zenodo.18501018}). The data is released under the CC BY-NC 4.0 license. The repository includes the raw image files organized by taxonomic family and comprehensive metadata provided in the COCO (Common Objects in Context) annotation format via JSON files, facilitating its immediate use in object detection and classification pipelines, as shown in Figure \ref{fig:flow_final_root}. The dataset does not include structured environmental metadata or detailed locality-level attributes, as its primary focus is on computer vision and object detection tasks.

\balance

\section*{INSIGHTS, LIMITATIONS AND USAGE NOTES}

The DAPWH dataset provides a multi-perspective visual record of Hymenoptera, featuring specimens captured in lateral, frontal, and dorsal views to ensure comprehensive morphological coverage. Although the current annotations and benchmarks are restricted to the family level, the dataset preserves higher-resolution taxonomic information in the image filenames. This enables straightforward extension to subfamily or genus level classification as additional annotations become available.

A significant subset of these images includes high-fidelity annotations of the insect body, wings, and a reference scale bar. These annotations facilitate the automated extraction of phenotypic features and biometric measurements, such as wing venation patterns. While current annotations focus on wing and body segmentation, the dataset's structure invites the scientific community to contribute additional labels for structures such as antennae or ovipositors, supporting future research in evolutionary biology.

The classification and detection benchmarks performed on this dataset demonstrate its robust utility for automated entomological research. The image-level classification models achieved high performance, with Top-1 accuracy reaching 92.28\% and YOLOv12 attaining an F1-score of 95.59\%. Similarly, the object detection benchmarks yielded strong results, particularly with YOLOv12, which achieved an mAP@50 of 90.53\%. These metrics validate the dataset's high quality and its effectiveness in supporting state-of-the-art computer vision architectures for the precise identification of Hymenoptera.

While the detection performance for the scale bar was lower than that of other classes, these objects remain high-value assets within the dataset. Each scale bar is accompanied by a precise metric annotation in millimeters (mm), ensuring that, even when automated detection requires refinement, the physical reference remains available for manual or semi-automated calibration.

\section*{SOURCE CODE AND SCRIPTS} 
The complete computational pipeline, including scripts for data preprocessing, dataset splitting, model training, and learning curve generation, is available in the GitHub repository: \url{https://github.com/joaomh/DAPWH-2026}.

\section*{ACKNOWLEDGEMENTS}
J.M.H.P., and G.N.H. contributed equally to this article, developed the dataset, annotated the images, drafted the manuscript, and performed technical validation. L.B.R.F., A.D.S., and H.C.O. reviewed the taxonomy of the insects, provided images of Ichneumonoidea and Vespidae and revised the manuscript. E.A.B.A. provided images of Apoidea families revised the manuscript and approved the final version. R.V.G., and M.A.C.V. performed technical validation, reviewed, revised, and approved the final version of the manuscript. A.M.P.D., and M.B. led the project and supervised the research, and reviewed, revised, and approved the final version of the manuscript.
    
This work was supported by the Petróleo Brasileiro S.A. - Petrobras, using resources from the R\&D clause of the ANP, in partnership with the Universidade de São Paulo (USP) and the Fundação de Apoio à Física e à Química (FAFQ), under Cooperation Agreement No. 2023/00016-6 and 2023/00013-7, Coordenação de Aperfeiçoamento de Pessoal de Nível Superior (CAPES) grant nº88887.002221/2024-00 and nº88887.975224/2024–00, Fundação de Amparo à Pesquisa do Estado de São Paulo (FAPESP) grant nº2014/50940-2, 2019/09215-6 and 2022/11451-2, Conselho Nacional de Desenvolvimento Científico e Tecnológico (CNPq) grant nº465562/2014-0, Instituto Nacional de Ciência e Tecnologia dos Hymenoptera Parasitoides (INCT-HYMPAR).

The authors have declared no conflicts of interest.
\bibliographystyle{IEEEtran}
\bibliography{references}
\end{document}